\documentclass{article}

\usepackage{microtype}
\usepackage{graphicx}
\usepackage{booktabs} 

\usepackage{url}
\usepackage{graphicx}
\usepackage{subfig}
\usepackage{multirow}
\usepackage{wrapfig}
\usepackage{xcolor,soul}
\usepackage{mathtools}
\usepackage{caption}
\usepackage{amsfonts}
\usepackage{fancyhdr}

\usepackage{hyperref}

\newif\ifcomment

\commenttrue

\ifcomment
\definecolor{stelios_colour}{RGB}{144, 238, 144}
\newcommand{\stelios}[1]{\sethlcolor{stelios_colour}\hl{[Stelios: #1]}}
\newcommand{\da}[1]{\sethlcolor{white}\hl{#1}}
\newcommand{\dafinal}[1]{\sethlcolor{green}\hl{#1}}
\else
\newcommand{\stelios}[1]{}
\newcommand{\da}[1]{}
\newcommand{\dafinal}[1]{}
\fi

\newcommand{\tb}{\color{black}} 
\newcommand{\rv}{\color{black}} 

\definecolor{stelios_colour}{RGB}{0, 0, 0} 
\newcommand{\steliosrev}[1]{\textcolor{stelios_colour}{#1}}

\definecolor{da_colour}{RGB}{0,0,0}
\newcommand{\darev}[1]{\textcolor{da_colour}{#1}}

\usepackage[accepted]{icml2020}

\icmltitlerunning{MuPPET: A precision-switching strategy for quantised fixed-point training of CNNs}

\begin{document}

\twocolumn[
\icmltitle{Multi-Precision Policy Enforced Training (MuPPET): \\ A precision-switching strategy for quantised fixed-point training of CNNs}



\icmlsetsymbol{equal}{*}

\begin{icmlauthorlist}
\icmlauthor{Aditya Rajagopal}{equal,ic}
\icmlauthor{Diederik Adriaan Vink}{equal,ic}
\icmlauthor{Stylianos I. Venieris}{samsung}
\icmlauthor{Christos-Savvas Bouganis}{ic}
\end{icmlauthorlist}

\icmlaffiliation{ic}{Intelligent Digital Systems Lab, Department of Electrical Engineering, Imperial College London}
\icmlaffiliation{samsung}{Samsung AI Center, Cambridge}

\icmlcorrespondingauthor{Aditya Rajagopal}{ar4414@ic.ac.uk}

\icmlkeywords{Machine Learning, ICML}

\vskip 0.4cm
]



\printAffiliationsAndNotice{\icmlEqualContribution} 

\thispagestyle{fancy}
\chead{PREPRINT: Accepted at the 37th International Conference on Machine Learning (ICML), 2020}
\renewcommand{\headrulewidth}{0pt}

\begin{abstract}
Large-scale convolutional neural networks (CNNs) suffer from very long training times, spanning from hours to weeks, limiting the productivity and experimentation of deep learning practitioners. 
As networks grow in size and complexity, training time can be reduced through low-precision data representations and computations.
However, in doing so the final accuracy suffers due to the problem of vanishing gradients.
Existing state-of-the-art methods combat this issue by means of a mixed-precision approach utilising two different precision levels, FP32 (32-bit floating-point) and FP16/FP8 (16-/8-bit floating-point), leveraging the hardware support of recent GPU architectures for FP16 operations to obtain performance gains.
This work pushes the boundary of quantised training by employing a multilevel optimisation approach that utilises multiple precisions including low-precision fixed-point representations.
The novel training strategy, MuPPET, combines the use of multiple number representation regimes together with a precision-switching mechanism that decides at run time the transition point between precision regimes. 
Overall, the proposed strategy tailors the training process to the hardware-level capabilities of the target hardware architecture and yields improvements in training time and energy efficiency compared to state-of-the-art approaches.
Applying MuPPET on the training of AlexNet, ResNet18 and GoogLeNet on ImageNet (ILSVRC12) and targeting an NVIDIA Turing GPU, MuPPET achieves the same accuracy as standard full-precision training with training-time speedup of up to 1.84$\times$ and an average speedup of 1.58$\times$ across the networks.
\end{abstract}

\section{Introduction}
\label{sec:introduction}

Convolutional neural networks (CNNs) have demonstrated unprecedented accuracy in various machine learning tasks, from video understanding \citep{Gan2015, He_2018} 
to drone navigation \citep{Loquercio_2018, kouris_drone2018iros}. 
To achieve such high levels of accuracy in inherently complex applications, current 
methodologies employ the design of large and complex CNN models \citep{Szegedy_2017, huang2017densely} trained over large datasets \citep{deng2009imagenet, coco2014}. 
Despite the associated accuracy gains, the combination of large models with massive datasets also results in excessively long training times.
This in turn leads to long turn-around times which set a limit to the productivity of deep learning practitioners and prohibit wider experimentation.
For instance, automatic tuning and search of neural architectures \citep{efficient_arch_search, block-wise_arch_search} is a rapidly advancing area where accelerated training enables the improvement of the produced networks.

To counteract these long turn-around times, substantial research effort has been invested in hyperparameter tuning for the acceleration of training, with a particular focus on batch size and content.
One line of work aims to maximise memory and hardware utilisation by changing the batch size, in order to perform CNN training-specific prefetching, scheduling and dependency improvement \citep{moDNN, vDNN}. 
Other works focus on altering batch size or reconstructing minibatches to improve the convergence rate while sustaining high hardware utilisation \citep{AdaBatch, RAIS, typicality_sampling}, demonstrating up to 6.25$\times$ training speedup.



In a different direction, a number of studies have focused on the use of reduced-precision training schemes. 
Reduced-precision arithmetic involves the utilisation of data formats that have smaller wordlengths than the conventional {\tb 32-bit floating-point (FP32)} representation and is an approach for co-optimising processing speed, memory footprint and communication overhead. 
{\tb Existing literature can be categorised into i) works that use reduced precision to only accelerate the training stage while eventually yielding an FP32 model, and ii) those that \darev{produce} networks with quantised weights.}
\darev{
Regarding the former, Courbariaux et al.~\yrcite{courbariaux2015training} and Gupta et al.~\yrcite{Gupta_2015} utilise dynamic quantisation and stochastic rounding respectively as a means to combat the accuracy loss due to quantisation.
}
%
Nevertheless, the effectiveness of the proposed schemes \darev{has only} been demonstrated on small-scale datasets {\tb such as CIFAR-10 and MNIST,} and on a limited set of networks.
Furthermore, the range of quantisation levels that has been explored varies greatly, with a number of works focusing solely on mild quantisation levels such as half-precision floating-point (FP16) \citep{micikevicius2018mixed}, {\tb while others focus on lower precisions such as 8-bit floating-point (FP8) \citep{8bitFPTrain}.}
{\tb Finally, quantisation has also been used as a means of reducing the memory and communication overhead in distributed training \citep{Sa2015, asyncsgd2017isca, Dan2017}.}


At the same time, the characteristics of modern CNN workloads and the trend towards quantised models have led to an emergence of specialised hardware processors, with support for low-precision arithmetic at the hardware level. 
From custom designs such as Google's TPUs \citep{Jouppi2017} and Microsoft's FPGA-based Brainwave system \citep{brainwave_2018} to commodity devices such as NVIDIA's Turing GPUs \cite{nvidia_turing_2019}, existing platforms offer increased parallelism through native support for reduced-precision data types including 16-bit floating-point (FP16), as well as 8- (INT8) and 4-bit (INT4) fixed-point. 
Although these platforms have been mainly designed for the inference stage, the low-precision hardware offers significant opportunities for accelerating the time-consuming training stage.
Consequently, there is an emerging need to provide training algorithms that can leverage these existing hardware optimisations and yield higher training speed.

This work tackles the field of reduced-precision training at an algorithmic level. 
Independently of the number of quantisation levels chosen, or how extreme the quantisation is, this work proposes a metric that estimates the amount of information each new training step obtains for a given quantisation level, \steliosrev{by capturing the diversity of the computed gradients \textit{across epochs}}. 
\darev{This enables the design of a policy that, \steliosrev{given a set of quantisation levels,} decides \textit{at run time} appropriate points to increase the precision of the training process \steliosrev{at that current instant} without impacting the achieved test accuracy compared to training in FP32. 
Due to its agnostic nature, it \steliosrev{remains} orthogonal and complementary to existing low-precision training schemes.}
\steliosrev{Furthermore, by pushing the precision below the 16-bit bitwidth of existing state-of-the-art techniques, the proposed method is able to leverage the low-precision capabilities of modern processing systems to yield training speedups without penalising the resulting accuracy, significantly improving the time-to-accuracy trade-off.}

\section{Background and Related Work}
\label{sec:related_works}
\label{subsec:mixed_prec}
The state-of-the-art method in training with reduced precision is mixed-precision training \citep{micikevicius2018mixed}.
The authors propose to employ low-precision FP16 computations in the training stage of high-precision CNNs that perform inference in FP32.
Along the training phase, the algorithm maintains a high-precision FP32 copy of the network's weights, known as a \textit{master copy}. 
At each minibatch, the inputs and weights are quantised to FP16 with all computations of the forward and backward pass performed in FP16, yielding memory footprint and runtime savings. 
Under this scheme, each stochastic gradient descent (SGD) update step entails accumulating FP16 gradients into the FP32 master copy of the weights, with this process performed iteratively throughout the training of the network. 
{\tb Micikevicius et al.~\yrcite{micikevicius2018mixed} evaluate their scheme over a set of state-of-the-art models on ImageNet, and show that mixed-precision training with FP16 computations achieves comparable accuracy to standard FP32 training.}


{\tb
Wang et al.~\yrcite{8bitFPTrain} also presented a method to train an FP32 model using 8-bit floating-point (FP8). 
The authors propose a hand-crafted FP8 data type, together with a chunk-based computation technique, and employ strategies such as stochastic rounding to alleviate the accuracy loss due to training at reduced precision. 
For AlexNet, ResNet18 and ResNet50 on ImageNet, Wang et al.~\yrcite{8bitFPTrain} demonstrate comparable accuracy to FP32 training while performing computations in FP8.
}

{\tb
Additionally, the works presented in \citep{Zhou_16, chen2017fxpnet} approach the problem of reduced-precision training through fixed-point computations.
\mbox{FxpNet}~\citep{chen2017fxpnet} was only evaluated on \mbox{CIFAR-10}, failing to demonstrate performance on more complex datasets such as ImageNet.
DoReFa-Net~\citep{Zhou_16} was tested on ImageNet, but only ran on AlexNet missing out on state-of-the-art networks such as GoogLeNet and ResNet.
}

All related works focus on either accelerating the training of an FP32 model through reduced-precision floating-point computations or using reduced-precision fixed-point computations to train a fixed-point model.
At the hardware level, 8-bit fixed-point multiplication uses 18.5$\times$ less energy and 27.5$\times$ less area with up to 4$\times$ lower runtimes than FP32 \citep{LowPrecSurvey,kouris2018cascadecnn}.
Consequently, this work attempts to push the boundaries of {\tb reduced-precision} training by combining the processing performance gains of low-bitwidth fixed-point computations with the floating-point-level accuracy of training an FP32 model.

\darev{Preliminary tests (Sec.~\ref{subsec:efficacy} for details) demonstrated that training solely in 8-bit fixed-point results in a significant degradation of validation accuracy compared to full FP32 training.} 
This work aims to counteract this degradation by \textit{progressively} increasing the precision of computations throughout training in an online manner determined by the proposed metric inspired by gradient diversity \citep{gradient_diversity}.
Additionally by operating in an online fashion, MuPPET tailors the training process to best suit 
the particular network-dataset pair at each stage of the training process.

{\tb
Gradient diversity was introduced by Yin et al. \yrcite{gradient_diversity} as a metric of 
{\tb measuring the dissimilarity} 
between sets of gradients that correspond to different minibatches. 
The gradient diversity of a set of gradients is defined as 
\begin{equation}
    \begin{split}
    \Delta_{\mathcal{S}}(\mathbf{w}) &= \frac{\sum_{i=1}^{n} ||\nabla f_{i}(\mathbf{w})||_{2}^{2}}{||\sum_{i=1}^{n}\nabla f_{i}(\mathbf{w})||_{2}^{2}}\\
    &= \frac{\sum_{i=1}^n||\nabla f_i(\mathbf{w})||_2^2}{\sum_{i=1}^n||\nabla f_i(\mathbf{w})||_2^2 + \sum_{i\ne j} \langle \nabla f_i(\mathbf{w}), \nabla f_j(\mathbf{w})\rangle}
    \end{split}
    \label{equ:gradient_div}
\end{equation}
where $\nabla f_i(\textbf{w})$ is the gradient of weights \textbf{w} for minibatch $i$. 

The key point to note in Eq.~(\ref{equ:gradient_div}) is that the denominator contains the inner product between two gradients from \textit{different minibatches}. 
Thus, orthogonal gradients would result in high gradient diversity, while similar gradients would yield low gradient diversity. 
The proposed framework, MuPPET, enhances this concept {\tb by considering} gradients \darev{between minibatches} \textit{across epochs} and proposes the developed metric as a proxy for the amount of new information gained in each training step. 
Section \ref{sec:implementation} further expands on how gradient diversity is incorporated into the MuPPET algorithm.
}



\section{Methodology - MuPPET Algorithm}
\label{sec:implementation}

\subsection{Multilevel Optimisation for Training CNNs}
\label{subsec:multilevel_opt}
Conventionally, the training process of a CNN can be expressed as in Eq. (\ref{equ:original_opt}). 
Given a CNN model $f$ parametrised by a set of weights $\textbf{w} \in \mathbb{R}^{D}$, {\tb where $D$ is the number of weights of $f$,} training involves a search for weight values that minimise the task-specific empirical loss, $Loss$, on the target dataset. Typically, a fixed arithmetic precision is employed across the training algorithm with FP32 currently being the \textit{de facto} representation used by the deep learning community.
\begin{equation}
    \min\limits_{\textbf{w}^{(\mathrm{FP32})} \in \mathbb{R}^{D}} {Loss}(f(\textbf{w}^{(\mathrm{FP32})}))
    \label{equ:original_opt}
\end{equation}
The proposed method follows a different approach by introducing a multilevel optimisation scheme \citep{migdalas2013multilevel} that leverages the performance gains of reduced-precision arithmetic. 
The single optimisation problem of Eq. (\ref{equ:original_opt}) is transformed into a series of optimisation problems with each one employing different precision for {\tb computations, but maintaining weights storage at FP32 precision.} 
Under this scheme, an $N$-level formulation comprises $N$ {\tb sequential} optimisation problems to be solved, with each level corresponding to a ``finer" model. 

Overall, this formulation adds a hierarchical structure to the training stage, with increasing arithmetic precision across the hierarchy of optimisation problems. 
Starting from the $N$-th problem, the inputs, weights and activations of the CNN model $f$ are quantised with precision $q^N$, which is the lowest precision in the system and represents the coarsest version of the model. 
Each of the $N$ levels progressively employs higher precision until the first level is reached, which corresponds to the original problem of \mbox{Eq. (\ref{equ:original_opt})}.
%
Formally, at the i-th level, the optimisation problem is formulated as
\begin{equation}
\begin{aligned}
    &\min\limits_{\textbf{w}^{(q^i)} \in \mathcal{V}} {Loss}(f(\textbf{w}^{(q^i)}))
    &\mathrm{s.t. } \mathcal{V} = \Big\{ \textbf{w}^{(q^i)} \in [\mathrm{LB}, \mathrm{UB}]^{D} \Big\} 
    \label{equ:low_prec_opt}
\end{aligned}
\end{equation}
%
\steliosrev{where LB and UB are the lower and upper bound in the representational range of precision $q^i$.}
The target CNN model $f$ uses a set of weights quantised with precision $q^i$ and hence the solution of this optimisation problem can be interpreted as an approximation to the original problem of \mbox{Eq. (\ref{equ:original_opt})}. To transition from one level to the next, the result of each level of optimisation is employed as a starting point for the next level, up to the final outermost optimisation that reduces to Eq. (\ref{equ:original_opt}).


\subsection{Quantised Training}
\begin{figure}[t!]
    \centering
        \includegraphics[trim={6cm 2cm 0cm 2cm},clip,width=0.7\textwidth]{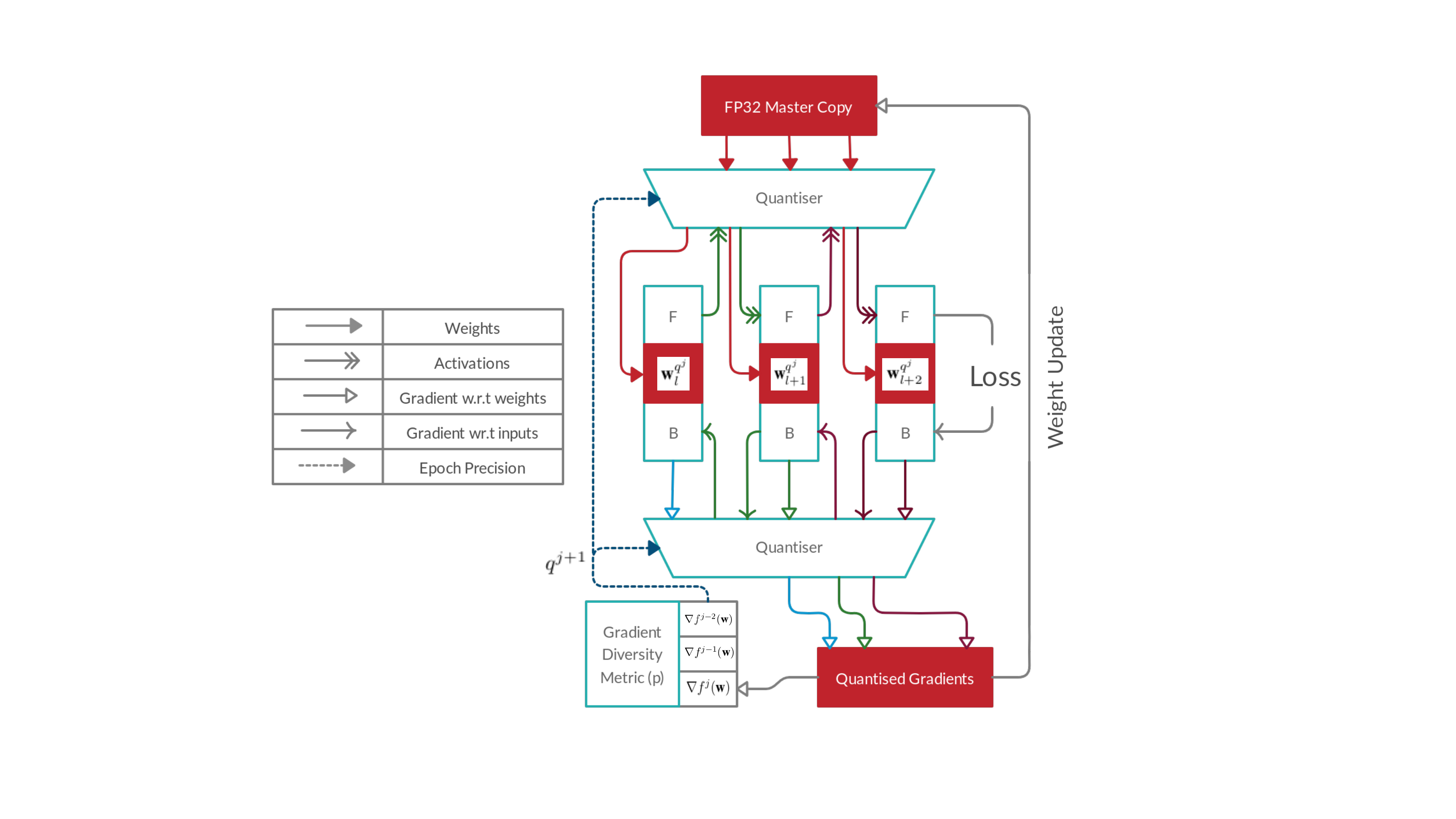}
    \caption{Quantised training scheme of MuPPET.}
    \label{fig:muppet_proc_diag}
\end{figure}
\label{subsec:algorithm}
%
Fig. \ref{fig:muppet_proc_diag} presents the process of training a CNN using {\tb the proposed algorithm}.
{\rv At epoch $j$}, MuPPET performs mixed-precision training where the weights are stored in an FP32 master copy and are quantised to the desired fixed-point precision {\rv ($q^j$)} on-the-fly. 
At epoch $j$, the computations for the forward and backward passes ($F$ and $B$ blocks respectively) are performed at the current quantised precision ($q^j$) and the activations as well as the gradients obtained from each layer are quantised by the quantiser module before being passed on to the next layer, or stored. 
After each minibatch, the full-precision master copy of the weights is updated using a quantised gradient matrix. 
As discussed in Section \ref{subsec:multilevel_opt}, the quantisation level is gradually increased over the period of the training. 
\da{To not compromise the final validation accuracy, switching between these optimisation levels at the correct times is crucial and controlled by a proposed precision-switching policy described in } Section \ref{subsec:prec_switch}.
\subsubsection{Quantisation Strategy}
\label{subsubsec:quant_strat}
In order to implement quantised training, a quantisation strategy needs to be defined. 
The proposed quantisation strategy utilises block floating-point arithmetic (also known as dynamic fixed-point), where each fixed-point number is represented as a pair of a $\mathrm{WL}^{\mathrm{net}}$-bit signed integer $x$ and a scale factor $s$, such that the value is represented as $x \times 2^{-s}$. 

During the forward and backward passes of the training process, the weights and feature maps are both quantised, and the multiplication operations are performed at the same low precision.
The quantisation method employs a stochastic rounding methodology \citep{Gupta_2015}.
The accumulation stage of the matrix-multiply operation is accumulated into a 32-bit fixed-point value to prevent overflow on the targeted networks.\footnote{The accumulator wordlength is large enough to accommodate the current CNN models without overflow.}
The result of this matrix multiplication is subsequently quantised to the target wordlength before being passed as input to the next layer. 
Following the block floating-point scheme, quantisation is performed such that each weight and feature map matrix in the network has a single scale factor shared by all values within the matrix. 
The quantisation configuration for the i-th level of optimisation and the l-th layer, $q_l^i$, and the full set of configurations, $q^i$, are given in Eq. (\ref{equ:quant_scheme}) respectively. 
\begin{equation}
    \begin{split}
    & q_l^i = \left< \mathrm{WL}^{\mathrm{net}}, s_{l}^{\mathrm{weights}}, s_{l}^{\mathrm{act}} \right>^i,\\ &~~~ \forall l \in \left[1,|\mathcal{L}|\right] ~~~~\mathrm{and}~~~~  q^i = \left< q_l^i ~|~ \forall l \in \left[1,|\mathcal{L}|\right] \right>
    \end{split}
    \label{equ:quant_scheme}
\end{equation}

%
where $|\mathcal{L}|$ is the number of layers of the target network, $\mathrm{WL}^{\mathrm{net}}$ is the fixed wordlength across the network, $s_{l}^{\mathrm{weights}}$ and $s_{l}^{\mathrm{act}}$ are the scaling factors for the weights and activations respectively, of the l-th layer for the i-th level of optimisation. 
As a result, for $N$ levels, there are $N$ distinct quantisation schemes; {\tb $N-1$ of these schemes are with varying fixed-point precisions and the finest level of quantisation, $q^1$, is single-precision floating-point (FP32). \steliosrev{The scaling factor for a matrix $\mathbf{X}$ is first calculated as shown in Eq.~(\ref{equ:scale_factor}) and individual elements are quantised as in Eq.~(\ref{equ:quant_value}).}}
\begin{equation}
    \begin{split}
    & s^{\mathrm{\{weights, act\}}} = \\ &\Bigl\lfloor \log_2 \left( \min \left(\frac{\mathrm{UB} + 0.5}{\mathbf{X}^{\mathrm{\{weights, act\}}}_{\mathrm{max}}}, \frac{\mathrm{LB} - 0.5}{\mathbf{X}^{\mathrm{\{weights, act\}}}_{\mathrm{min}}} \right) \right) \Bigl\rfloor
    \end{split}
    \label{equ:scale_factor}
\end{equation}
\begin{equation}
    \begin{split}
    &x^{\mathrm{\{weights, act\}}}_{\mathrm{quant}} =\\ &\left\lfloor x^{\mathrm{\{weights, act\}}} \cdot 2^{s^{\mathrm{\{weights, act\}}}} + \mathrm{Unif}\left( -0.5, 0.5 \right) \right\rceil
    \end{split}
    \label{equ:quant_value}
\end{equation}
\steliosrev{where $\mathbf{X}^{\mathrm{\{weights, act\}}}_{\mathrm{\{max, min\}}}$ is either the maximum or minimum value in the weights or feature maps matrix of the current layer, LB and UB are the lower and upper bound of the current wordlength $\mathrm{WL}^{\mathrm{net}}$, and Unif(a,b) represents sampling from the uniform distribution in the range [a,b].}
\darev{Eq.~(\ref{equ:scale_factor}) adds $0.5$ and $-0.5$ to UB and LB respectively to ensure maximum utilisation of $\mathrm{WL}^{\mathrm{net}}$.}

\subsubsection{Information Transfer between Levels}
\label{subsubsec:info_transfer}
Employing multilevel training for CNNs requires an appropriate mechanism for transferring information between levels. 
To achieve this, the proposed optimiser maintains a master copy of the weights in full precision (FP32) throughout the optimisation levels. 
{\tb Similar to} mixed-precision training \citep{micikevicius2018mixed}, at each level the SGD update step is performed by accumulating a fixed-point gradient value into the FP32 master copy of the weights.
{\tb Starting from the coarsest quantisation level $i = N$}, to transfer the solution from level $i$ to level $i-1$, the master copy is quantised using the quantisation scheme $q^{i-1}$. 
With this approach, the weights are maintained in FP32 and are quantised on-the-fly during run time in order to be utilised in each training step. 

\subsection{Precision-Switching Policy}
\label{subsec:prec_switch}
MuPPET computes $\Delta_{\mathcal{S}}(\textbf{w})$ (Eq. (\ref{equ:gradient_div})) between gradients obtained across epochs as a proxy to measure the information that is obtained during the training process; the lower the diversity between the gradients, the less information this level of quantisation provides {\tb towards the training of the model.}  
Therefore, the proposed method comprises a novel normalised {\tb inter-epoch} version of the gradient diversity along with a run-time policy to determine the epochs in which to switch precision. 

The following policy is employed to determine when a precision switch is to be performed. 
{\tb For a network with layers $\mathcal{L}$ and a quantisation scheme $q^i$ that was switched into at epoch $e$:}
\begin{enumerate}
    \item For each epoch $j$ and each layer $l \in \mathcal{L}$, the last \darev{minibatch's} gradient, $\nabla f_{l}^{j}(\textbf{w})$, is stored.
    
    \item After $r$ (resolution) number of epochs, the {\tb inter-epoch} gradient diversity at epoch $j$ is
    \begin{equation}
        \Delta_{\mathcal{S}}(\textbf{w})^j = \frac{ \sum_{\forall l \in \mathcal{L}}{\frac{\sum_{k = j-r}^j ||\nabla f_l^k(\textbf{w})||_2^2}{|| \sum_{k=j-r}^j \nabla f_l^k(\textbf{w})||_2^2}}}{|\mathcal{L}|}
        \label{eq:interepoch_grad_div}
    \end{equation}
    \item At an epoch $j$, given a set of gradient diversities $\mathcal{S}(j) = \Big\{ \Delta_{\mathcal{S}}(\textbf{w})^i \ \ \forall \  e \le i < j \Big\}$, \\ the ratio \mbox{$p = \frac{\max \mathcal{S}(j)}{\Delta_{\mathcal{S}}(\textbf{w})^j}$}
    is calculated.
    
    
    \item An empirically determined decaying threshold  
    \mbox{$T = \alpha + \beta e^{-\lambda j} ~\refstepcounter{equation}(\theequation)\label{equ:decay_thresh}$}
    is placed on the ratio $p$.
        
    
    \item If $p$ violates $T$ more than $\gamma$ times, a precision switch is triggered and $\mathcal{S}(j) = \emptyset$. 
\end{enumerate}


As long as the gradients across epochs remain diverse, $\Delta_{\mathcal{S}}(\textbf{w})^j$ (Eq.(\ref{eq:interepoch_grad_div})) at the denominator of $p$ sustains a high value and the value of $p$ remains low. 
However, when the gradients across epochs become similar, $\Delta_{\mathcal{S}}(\textbf{w})^j$ decreases and the value of $p$ becomes larger.
\textit{Generalisability across epochs} is obtained as $p$ accounts for the change in information relative to the maximum information available since the last precision change. 
Hence, the metric acknowledges the presence of temporal variations in information provided by the gradients. 
\textit{Generalisability across networks} and \textit{datasets} is maintained as $p$ measures a ratio. 
Consequently, the absolute values of gradients which could vary between networks and datasets, matter less.
Overall, MuPPET employs the metric $p$ as a mechanism to trigger a precision switch whenever $p$ violates threshold $T$ more than $\gamma$ times.\\\\
The likelihood of observing $r$ gradients across $r$ epochs that have low gradient diversity, especially at early stages of training is low.
The intuition applied here is that when this does happen at a given precision, it may be an indication that information is being lost due to quantisation and thus corresponds to a high $p$ value, which argues to move to a higher bitwidth.

\subsubsection{Hyperparameters}
\label{subsubsec:hyperparam}


The hyperparameters for the proposed MuPPET algorithm are the following: 
1) values of $\alpha$, $\beta$, and $\lambda$ that define the decaying threshold from Eq. (\ref{equ:decay_thresh}), 2) the number of threshold violations allowed before the precision change is triggered ($\gamma$), 3) the resolution $r$, 4) the set of precisions at which training is performed, and 5) the epochs at which the learning rate is changed.
The values of $\alpha$, $\beta$, $\lambda$, $r$, and $\gamma$ were set at 1, 1.5, 0.1, 3, and 2 respectively after empirical cross-validation. 
These were tuned by running training on AlexNet and ResNet20 on the CIFAR-10 dataset. 
All MuPPET hyperparameters remain the same regardless of network or dataset.
Regarding training hyperparameters, batch size was increased from 128 to 256 going from CIFAR-10 to ImageNet.
All other training hyperparameters, {\tb including learning rate}, remained constant.
Analysis of generalisability and the training hyperparameters used are presented in Section \ref{subsec:generalisability}. 
The \darev{empirically-chosen quantised precisions} at which training was performed were 8-, 12-, 14- and 16-bit fixed-point.
Precisions below this did not result in any progress towards convergence for any network. 


Overall, MuPPET introduces a policy that decides at run time an appropriate point to switch between quantisation levels. 
After training at 16-bit fixed-point, the rest of the training is performed at FP32 until the desired validation accuracy is reached. 
Decaying the learning rate causes a finer exploration of the optimisation space as does increasing the quantisation level. 
Therefore, the learning rate was kept constant during quantised training and was decayed only after switching to FP32. 


\section{Evaluation of MuPPET}
\label{subsec:prec_switch_policy_eval}


%

\subsection{Generalisability}
\label{subsec:generalisability}
 \begin{figure*}[t!]
     \centering
     \begin{tabular}{ccc}
         \subfloat[AlexNet - CIFAR-10]{\includegraphics[trim=6 4 0 93,clip,width = 0.33\textwidth]{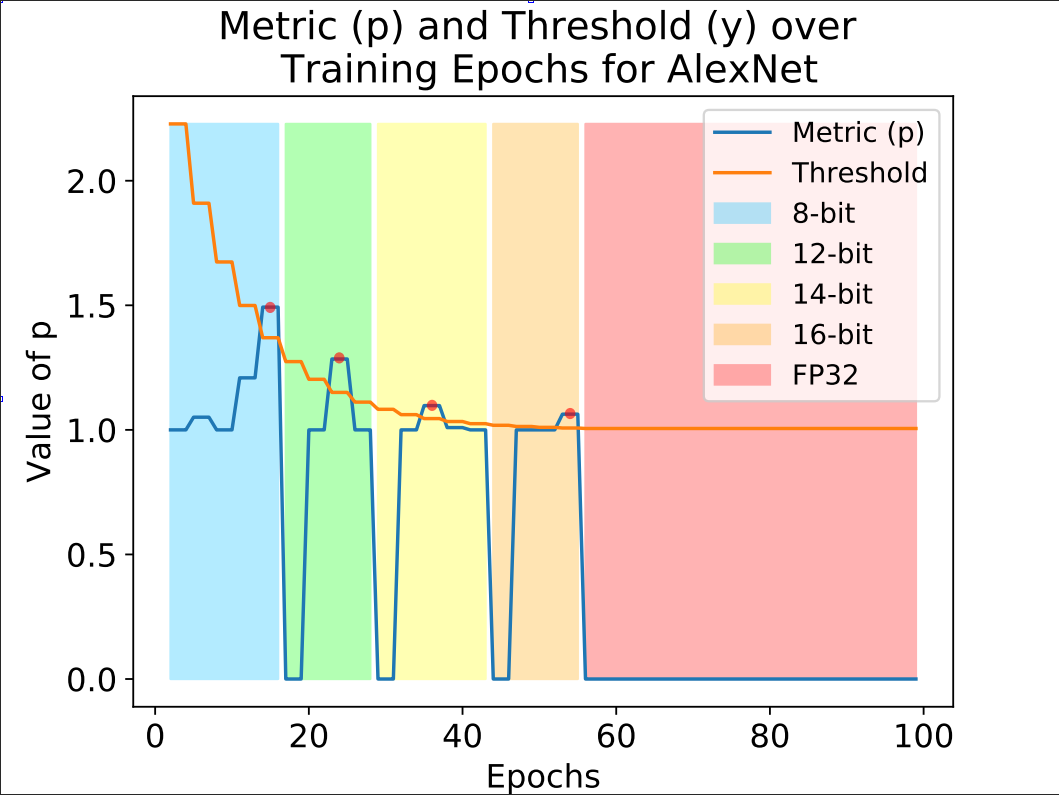} \label{fig:p_alexnet_cifar10}}
         \subfloat[ResNet20 - CIFAR-10]{\includegraphics[trim=6 5 0 98,clip,width = 0.33\textwidth]{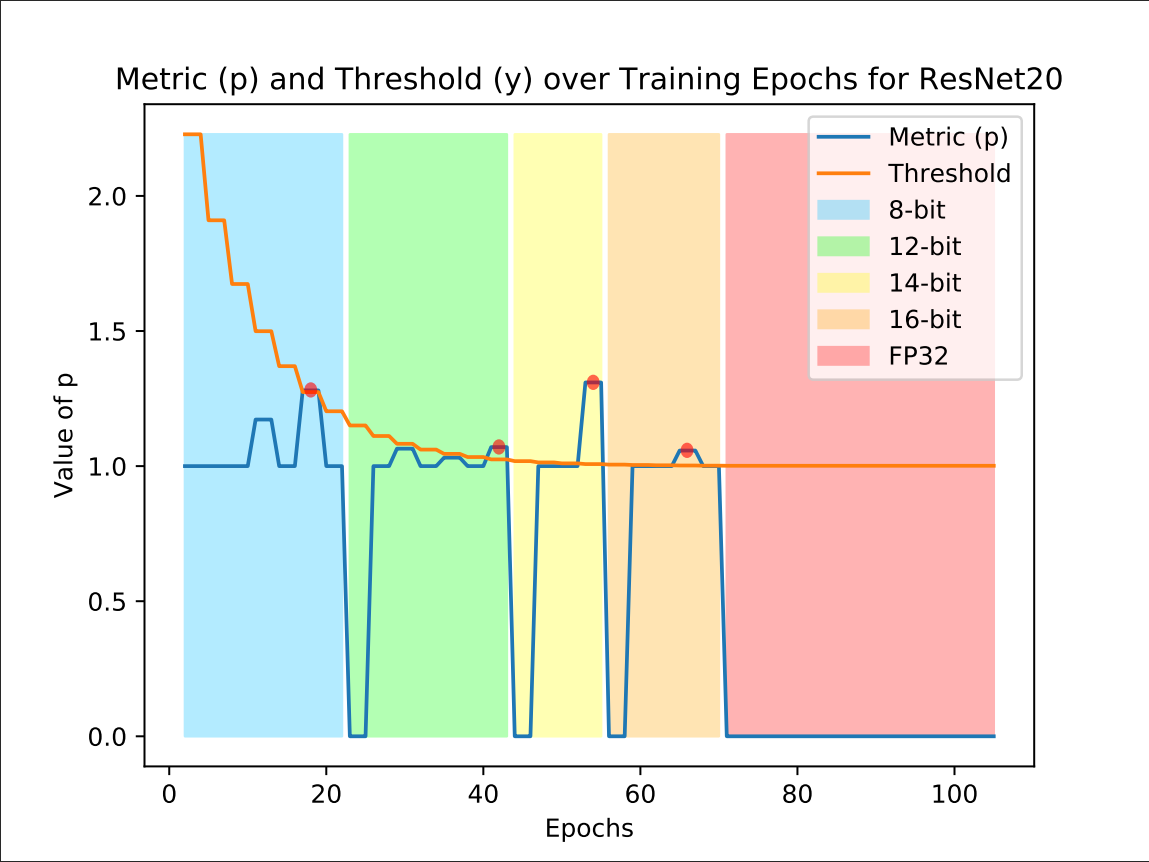} \label{fig:p_resnet_cifar10}} 
         \subfloat[GoogLeNet - CIFAR-10]{\includegraphics[trim=6 5 4 98,clip,width = 0.33\textwidth]{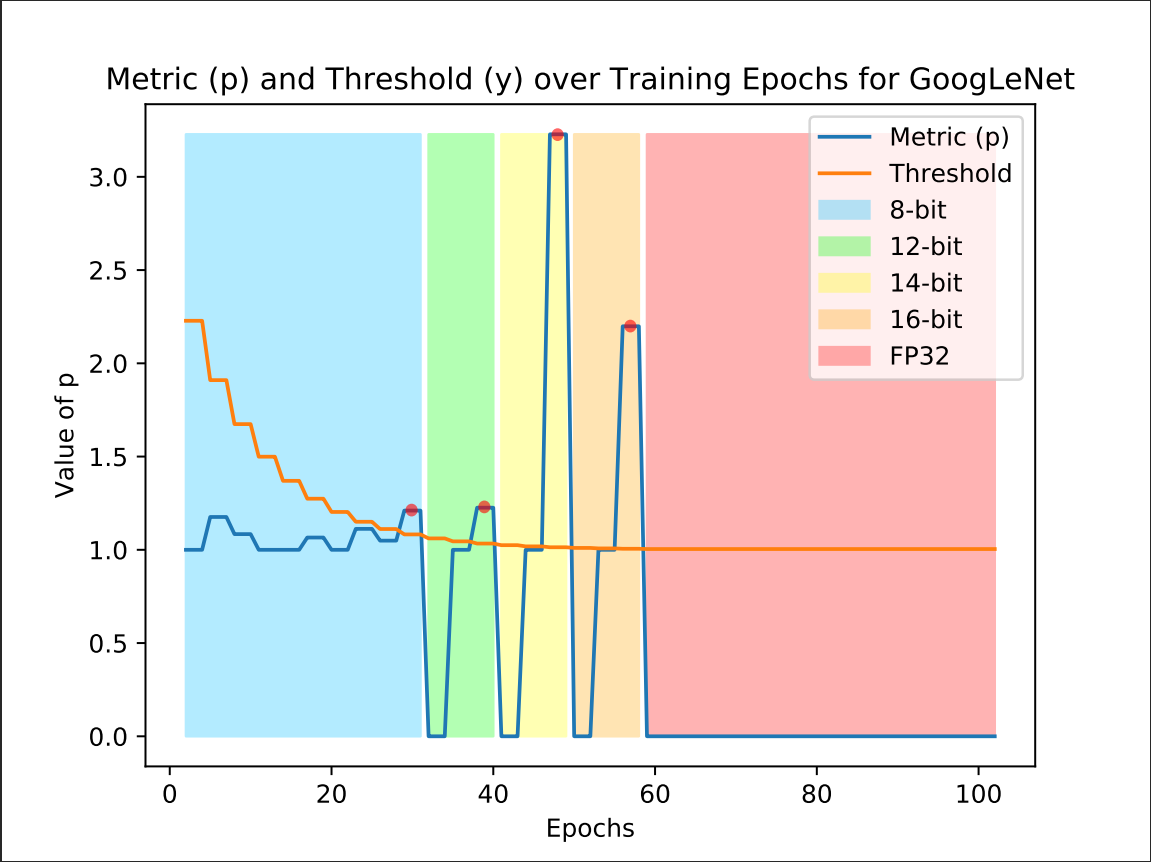} \label{fig:p_googlenet_cifar10}} \\
         \subfloat[AlexNet - ImageNet]{\includegraphics[trim=6 5 4 85,clip,width = 0.33\textwidth]{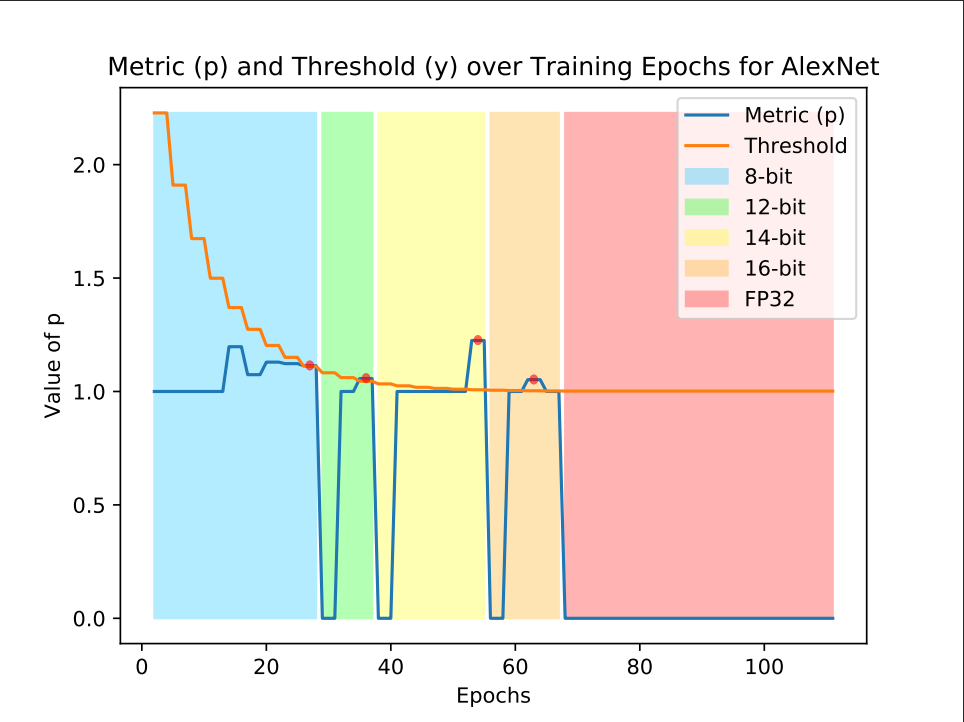} \label{fig:p_alexnet_imagenet}} 
         \subfloat[ResNet18 - ImageNet]{\includegraphics[trim=6 5 4 98,clip,width = 0.33\textwidth]{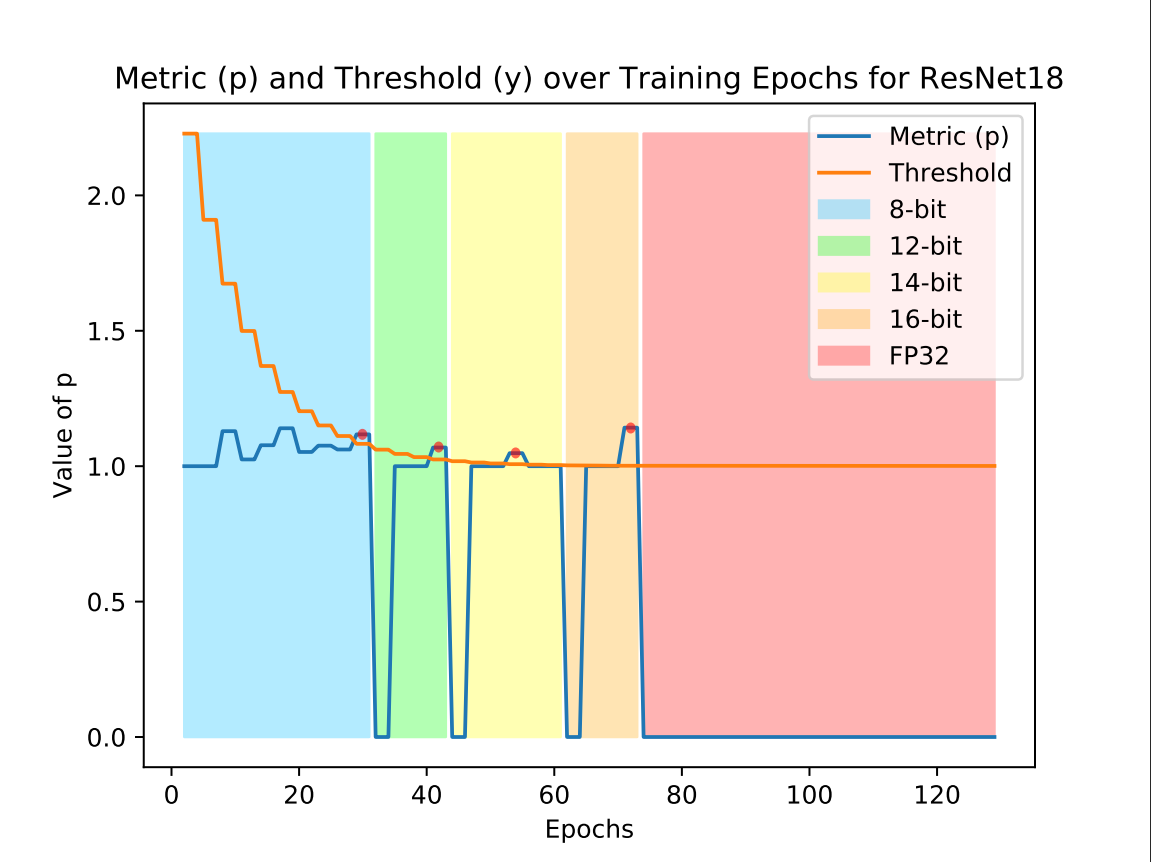} \label{fig:p_resnet_imagenet}}
         \subfloat[GoogLeNet - ImageNet]{\includegraphics[trim=6 5 0 98,clip,width = 0.33\textwidth]{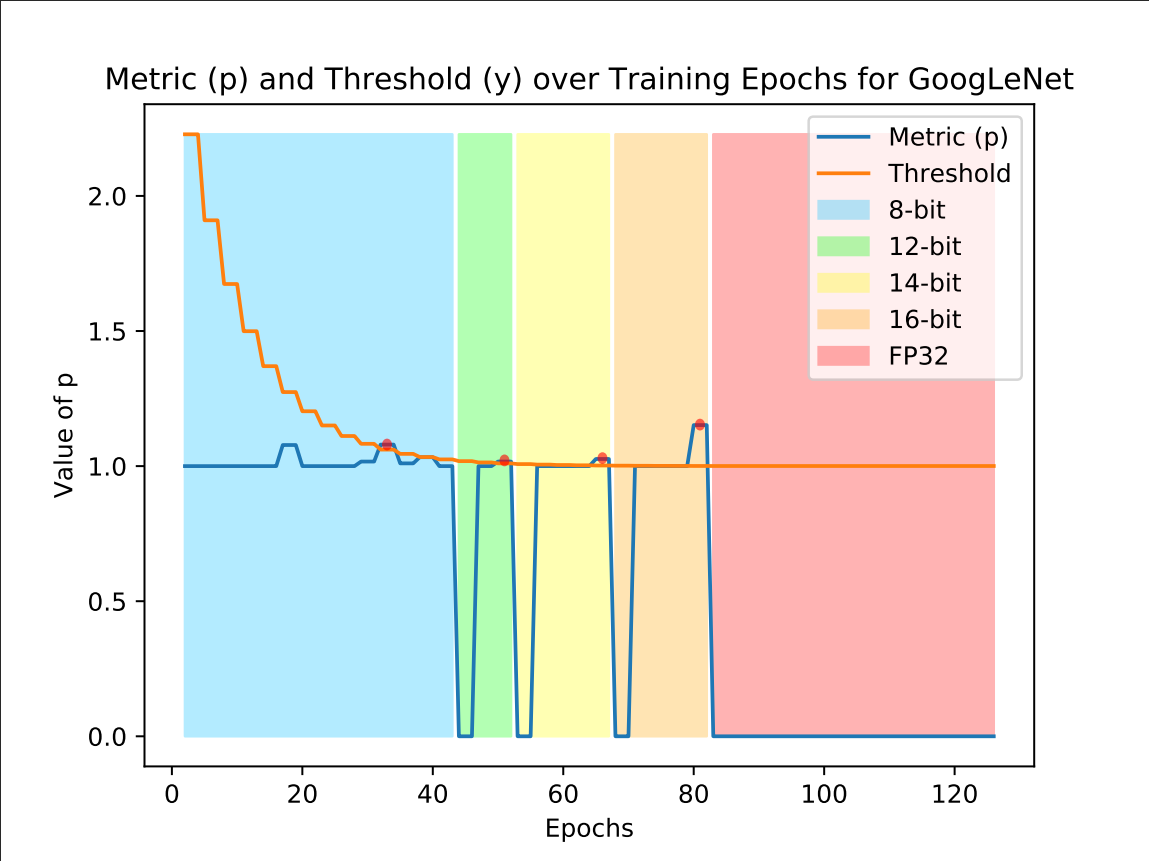} \label{fig:p_googlenet_imagenet}} 
     \end{tabular}
     \caption{Demonstration of the generalisability of $p$ over networks, datasets and epochs. Figures show the metric $p$ (blue) and the decaying threshold $T$ (orange) over epochs of training.}
     \label{fig:gd_metric}
 \end{figure*}
The MuPPET framework was evaluated on its applicability across epochs, networks and datasets. 
Fig. \ref{fig:gd_metric} shows the value of the metric $p$ over the epochs in blue, and the decaying threshold described in Eq. (\ref{equ:decay_thresh}) in orange. 
The number of epochs for which training in each precision was performed is shown by the various overlay colours.
The first violation is denoted by a red dot and the second violation is not seen as it occurs exactly at the point of switching. 
The graphs show that across various networks and datasets, the values of $p$ stay relatively similar, backing the choice of a universal decaying factor. 
{\tb Furthermore, empirical results for CIFAR-10 indicated that changing from one fixed-point precision to another too early in the training process had a negative impact on the final validation accuracy.} 
Using a decaying threshold ensures that the value of $p$ needs to be much higher in the initial epochs to trigger a precision change due to the volatility of $p$ in early epochs of training.


\subsection{Performance Evaluation}
\label{sec:acc_results}
The accuracy results presented in this section utilised the proposed stochastic quantisation strategy.
The methodology was developed using PyTorch. 
As the framework does not natively support low-precision implementations, all quantisation and computations corresponding to 8-, 12-, 14-, and 16-bit precisions were performed through emulation on floating-point hardware \footnote{\href{https://github.com/ICIdsl/pytorch\_training.git}{https://github.com/ICIdsl/pytorch\_training.git}}.
All hyperparameters not specified below were left as PyTorch defaults. 
For all networks, an SGD optimiser was used with batch sizes 128 on CIFAR-10/100 or 256 on ImageNet, momentum of 0.9 and weight decay \mbox{of 1$e^{-4}$.}

\begin{figure*}[t!]
    \centering
    \begin{tabular}{ccc}
        \subfloat[AlexNet]{\includegraphics[trim=0 0 0 40,clip,width = 0.33\textwidth]{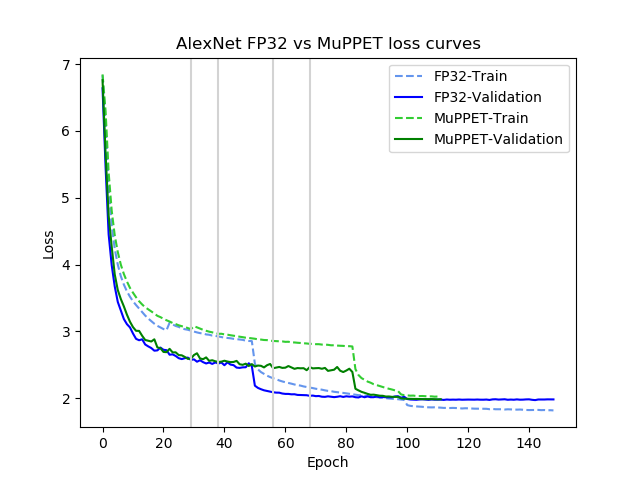} \label{fig:alexnet_train_test_loss}}
        \subfloat[ResNet18]{\includegraphics[trim=0 0 0 40,clip,width = 0.33\textwidth]{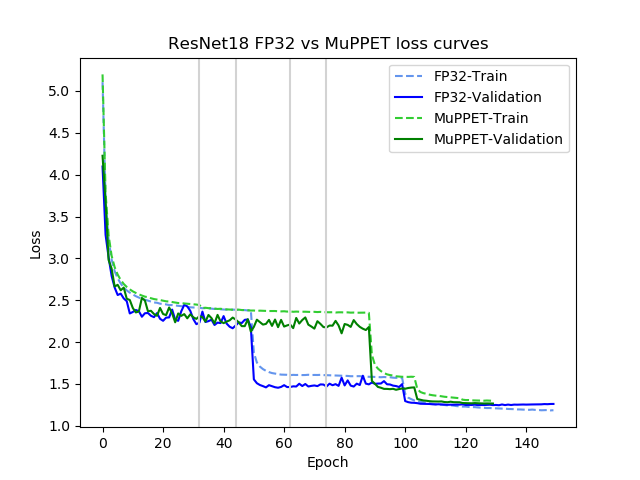} \label{fig:resnet18_train_test_loss}}
        \subfloat[GoogLeNet]{\includegraphics[trim=0 0 0 40,clip,width = 0.33\textwidth]{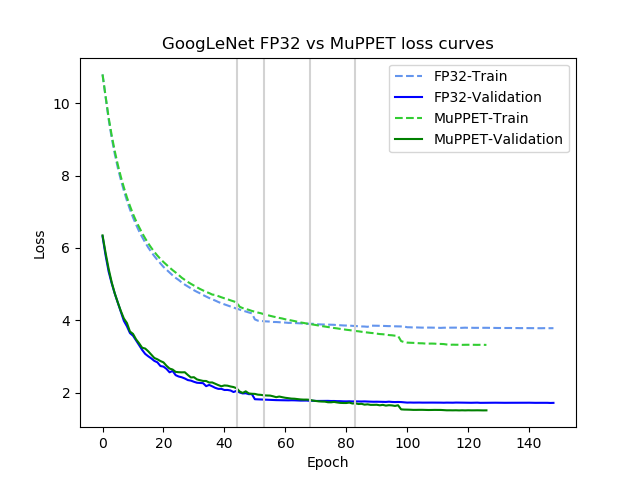} \label{fig:googlenet_train_test_loss}}
    \end{tabular}
    \caption{FP32 vs MuPPET training and validation loss graphs for different networks on the ILSVRC12 dataset.}
    \label{fig:imgnet_train_val_loss}
\end{figure*}

{\tb As a baseline,} an FP32 model with identical hyperparameters (except for batch size) was trained.
The baseline FP32 training was performed by training for 150 epochs and reducing the learning rate by a factor of 10 at epochs 50 and 100. 
In order to achieve comparable final validation accuracy to the FP32 baseline, once MuPPET triggered a precision change out of 16-bit fixed-point, 45 training epochs at FP32 precision were performed. 
The learning rate was reduced by a factor of 10 every 15 FP32 training epochs. 
For AlexNet, ResNet18, ResNet20, and GoogLeNet, the initial learning rate was set to 0.01, 0.1, 0.1, and 0.001 respectively. 
The ImageNet training and validation loss curves can be seen in the Fig.\ref{fig:imgnet_train_val_loss}.
For each of the graphs, the light grey lines indicate the epoch at which precision was switched in the MuPPET run.
The green lines follow MuPPET runs and the blue FP32 training. 
Solid lines show validation loss and dashed training loss.

\setlength{\tabcolsep}{2pt}
\begin{table}[h]
    \centering
    \caption{Top-1 test accuracy (\%) on CIFAR-10/100 and \mbox{ImageNet} (ILSVRC12 Validation Set) for FP32 baseline and MuPPET.}
    \vspace{0.2cm}
    \resizebox{1.0\columnwidth}{!}{
    \begin{tabular}{l c c c c c c c c c}
        \toprule
        \multirow{2}{*}{} & \multicolumn{3}{c}{\textbf{CIFAR-10}} & \multicolumn{3}{c}{\textbf{CIFAR-100}} & \multicolumn{3}{c}{\textbf{ImageNet}} \\ 
                      & FP32          & MuPPET & Diff (pp) & FP32          & MuPPET & Diff (pp)     & FP32          & MuPPET & Diff 
                      (pp) \\ \midrule
        \textbf{AlexNet}           & 75.45         & 74.49 & -0.96 &39.20 &38.19 & -0.99      & 56.21         & 55.33  & -0.88        \\ 
        \textbf{ResNet}            & 90.08         & 90.86 &  \phantom{-}0.78  &64.60 &65.80 &1.20     & 69.48         & 69.09 & -0.39        \\ 
        \textbf{GoogLeNet}         & 89.23         & 89.47  & \phantom{-}0.24 &62.90 &65.70 &2.80      & 59.15         & 63.70 & \phantom{-}4.55      \\ 
        \bottomrule
    \end{tabular}
    }
    \label{tab:Accuracies}
\end{table}

%
%

Table \ref{tab:Accuracies} presents the achieved Top-1 validation accuracy of MuPPET and the FP32 baseline, together with the accuracy difference in percentage points (pp). As shown on the table, MuPPET is able to provide comparable Top-1 validation accuracy to standard FP32 training across both networks and datasets. {\tb Due to a sub-optimal training setup of GoogLeNet on ImageNet, the baseline and MuPPET training severely underperformed compared to the reported state-of-the-art works. 
Nevertheless, the results demonstrate the quality of training with MuPPET using identical hyperparameters.}
\darev{As a result, MuPPET's performance demonstrates the effectiveness of the precision-switching strategy in achieving significant acceleration of training time (Section \ref{sec:perf_results}) at negligible cost in accuracy by running many epochs at lower precision, particularly on very large datasets.}



\subsection{Wall-clock Time Improvements}
\label{sec:perf_results}

\darev{This section explores the gains in estimated wall-clock time of the current implementation of MuPPET (Current Impl.) with respect to baseline FP32 training, Mixed Precision by Micikevicius et al.~\yrcite{micikevicius2018mixed} and MuPPET's ideal implementation (Table \ref{tab:wall_clock_times}).
}
For all performance results, the target platform was an NVIDIA RTX 2080 Ti GPU.

At the moment, deep learning frameworks, such as PyTorch, do not provide native support for reduced-precision hardware {\rv except for FP16 used by Micikevicius et al.~\yrcite{micikevicius2018mixed}}.
Consequently, the wall-clock times 
on Table \ref{tab:wall_clock_times} were estimated using a performance model developed with NVIDIA's CUTLASS library \citep{cutlass} for reduced-precision general matrix-multiplication (GEMM) employing the latest Turing architecture GPUs. 
{\rv The entire training process in PyTorch was profiled including the following MuPPET specific operations: quantisation of weights, activations and gradients as well as all calculations pertaining to $\Delta_{\mathcal{S}}(\mathbf{w})^j$.}
The GEMMs that were accelerated were in the convolutional and fully-connected layers of each network.
INT8 hardware was used to profile the 8-bit fixed-point computations, while FP16 hardware was used to profile 12-, 14-, and 16-bit fixed-point computations as well as Mixed Precision \citep{micikevicius2018mixed} wall-clock time.
CUTLASS \citep{cutlass} natively implements bit-packing to capitalise on improved memory-bandwidth utilisation.
{\rv To exploit further performance benefits by utilising 12- and 14-bit fixed-point computations, custom hardware such as FPGAs or ASICs would be required.}
{\rv The model for the current implementation is limited by the fact that current frameworks force quantisation of activations inbetween layers to happen to and from FP32. 
For the MuPPET (Ideal) scenario and Mixed Precision \cite{micikevicius2018mixed}, the model assumes native hardware utilisation for all deployed data representations which would eliminate the current overheads.} 

As shown on Table \ref{tab:wall_clock_times}, MuPPET consistently achieves \mbox{1.25-1.32$\times$} speedup over the FP32 baseline across the networks when targeting ImageNet on the given GPU.
With respect to Mixed Precision, the proposed method outperforms it on AlexNet by 1.23$\times$ and delivers comparable performance for ResNet18 and GoogLeNet. 
Currently, the absence of native quantisation support, and hence the necessity to emulate quantisation and the associated overheads, is the limiting factor for MuPPET to achieve higher processing speed. 
In this respect, \darev{MuPPET run on native hardware} would yield 1.05$\times$ and 1.48$\times$ speedup for ResNet18 and GoogLeNet respectively compared to Mixed Precision.
\steliosrev{As a result, MuPPET demonstrates consistently faster time-to-accuracy \citep{time_to_accuracy_2019} compared to Mixed Precision across the benchmarks. Additionally, while Mixed Precision has already reached its limit by using FP16 on FP16-native GPUs, the \darev{8-}, 12-, 14- and 16-bit fixed-point computations enabled by MuPPET leave space for further potential speedup when targeting next- and current-generation \citep{brainwave_2018} precision-optimised fixed-point platforms.}
{\tb
Similar to the analysis 
in Section \ref{sec:acc_results}, Micikevicius et al.~\yrcite{micikevicius2018mixed} and Wang et al.~\yrcite{8bitFPTrain} compare their 
schemes to baseline FP32 training performed by them.
The reported results demonstrate that their methods achieve similar accuracy results to our method by lying close to the respective FP32 training accuracy.
{\rv As Wang et al.~\yrcite{8bitFPTrain} do not provide any results for gains in wall-clock times, propose a custom FP8 data type and computation, provide emulated results from an in-house (not open source) machine learning framework, and could build only a subset of their proposed ideas on an ASIC, their work could not be directly compared to our method.}
}

\begin{table}[t]
    \centering
    \caption{Wall-clock time (GPU hours:mins) \& relative acceleration for networks targeting ImageNet.}
    \vspace{0.2cm}
    \resizebox{1\columnwidth}{!}{
    \begin{tabular}{lcccc}
        \toprule
        & \textbf{\begin{tabular}[c]{@{}c@{}}FP32\\ (Baseline)\end{tabular}} & \textbf{\begin{tabular}[c]{@{}c@{}}Mixed Prec\\ \citep{micikevicius2018mixed}\end{tabular}} & \textbf{\begin{tabular}[c]{@{}c@{}}MuPPET\\ (\darev{Current Impl.})\end{tabular}} & \textbf{\begin{tabular}[c]{@{}c@{}}MuPPET\\ (\darev{Ideal})\end{tabular}} \\ 
        
        \midrule
        \textbf{AlexNet}   & \phantom{1}30:13 (1$\times$)    & \phantom{1}29.20 (1.03$\times$)       & \phantom{1}23:52 (1.27$\times$)   & 20:25 (1.48$\times$)                                                                    \\
        \textbf{ResNet18}  & 132:46 (1$\times$)   & \phantom{1}97:25 (1.36$\times$)       & 100:19 (1.32$\times$)  & 92:43 (1.43$\times$)                                                                    \\
        \textbf{GoogLeNet} & 152:28 (1$\times$)   & 122:51 (1.24$\times$)      & 122:13 (1.25$\times$)  & 82:38 (1.84$\times$)                                                                   \\ 
        \bottomrule
        
    \end{tabular}
    }
    \label{tab:wall_clock_times}
\end{table}

\subsection{Precision Switching}
\label{subsec:efficacy}
\begin{figure}[t]
    \centering
    \includegraphics[trim=0 0 0 40 ,clip,width=0.5\textwidth]{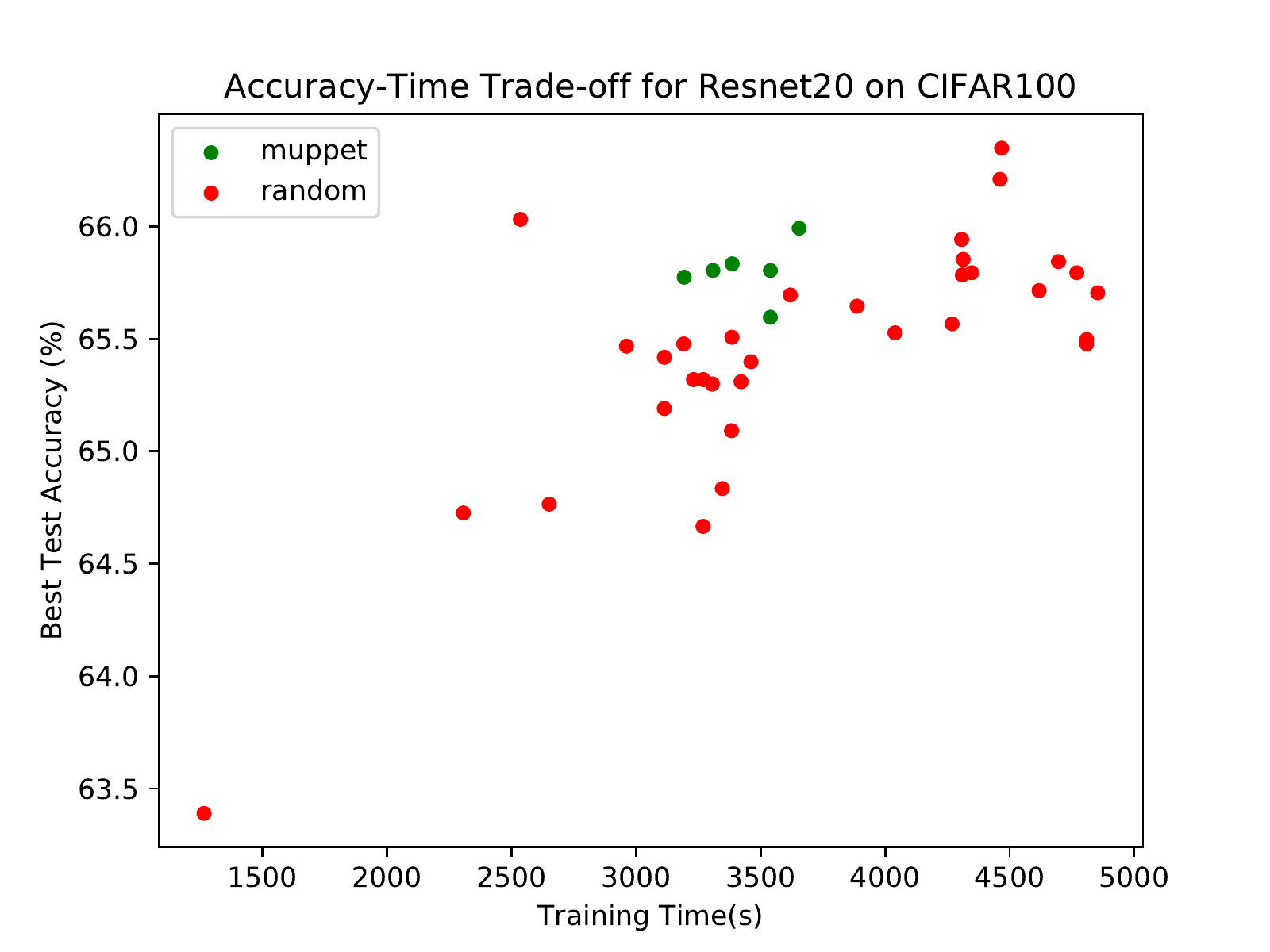}
    \vspace{-0.8cm}
    \caption{\centering Test accuracy vs time trade-off for ResNet20 MuPPET runs on CIFAR-100.}
    \label{fig:time_acc_graph}
\end{figure}


To evaluate the ability of MuPPET to effectively choose an epoch to switch precision at, AlexNet and ResNet20 were first trained using MuPPET on the CIFAR-100 dataset. 
{\tb The hyperparameters for MuPPET were kept the same across all runs.}
%
%
\darev{From the results it was noted that training at reduced precision and not switching at all causes a drop in validation accuracy of 1.4\% and 1.3\% for AlexNet and ResNet20 respectively}, hence demonstrating the need to switch precisions when training at bitwidths as low as 8-bit fixed-point.

\begin{table}[t]
\centering
\caption{Evaluation of MuPPET's ability to tailor to dataset and network combination using Top-1 test accuracy on CIFAR-100.}
\vspace{0.2cm}
\resizebox{0.95\columnwidth}{!}{
    \begin{tabular}{lc c c c}
    \toprule
    \multirow{2}{*}{} & \multicolumn{2}{c}{\textbf{ResNet20}} & \multicolumn{2}{c}{\textbf{GoogLeNet}} \\ 
     & CIFAR-10 & CIFAR-100 & CIFAR-10 & CIFAR-100 \\ \midrule
    \textbf{ResNet20} & 65.01 & \textbf{65.80} & - & 65.0 \\ 
    \textbf{GoogLeNet} & - & 64.00 &64.70  & \textbf{65.70} \\ 
    \bottomrule
    \end{tabular}
}
\label{tab:muppet_tailoring}
\vspace{-0.5cm}
\end{table}

{\rv
To demonstrate the benefits of a precision-switching methodology, two further sets of experiments were conducted on ResNet20 using CIFAR-100.
First, 34 training runs were performed (34 red dots in Fig. \ref{fig:time_acc_graph}), where for each run four epochs along the standard training duration were randomly selected and used as the switching points.
Fig. \ref{fig:time_acc_graph} shows the best test accuracy achieved by each of the runs and the training time as estimated by our performance model described in Section \ref{sec:perf_results}. 
It shows that for a given time-budget, MuPPET runs (6 green dots) outperform on average all other experiment sets, demonstrating the need for a precision-switching policy that is real-time in order to achieve a good accuracy-to-training-time trade-off.
The accuracy and time have mean $\pm$ standard deviation of (65.54\%$\pm$0.06, 3429.19s$\pm$46.1) and (65.44\%$\pm$0.09, 3679.15s$\pm$141.1) for the green and red points respectively. 

To demonstrate that MuPPET tailors the training process to the network-dataset pair while using a policy that is agnostic of the network-dataset combination, the experiments in Table \ref{tab:muppet_tailoring} were performed. 
The switching strategy generated by MuPPET for both ResNet20 and GoogLeNet on CIFAR-10 and CIFAR-100 were recorded.
The experiments shown in Table~\ref{tab:muppet_tailoring} used the CIFAR-100 dataset. 
The bold values show the accuracy achieved by the schedule proposed by MuPPET for the corresponding network-dataset pair.
For the ResNet20 row, column 1 shows the accuracy achieved if the precision schedule proposed by MuPPET on the CIFAR-10 dataset for ResNet20 was applied to CIFAR-100 training (dataset tailoring).
In the same row, column 4 shows the accuracy obtained when CIFAR-100 training was performed on ResNet20 using GoogLeNet's CIFAR-100 MuPPET schedule (network tailoring).
Similar results are shown for GoogLeNet.
As all experiments underperform compared to the MuPPET training process, Table~\ref{tab:muppet_tailoring} demonstrates the ability of MuPPET to tailor the process to a network-dataset combination.
}

\section{Conclusion}
\label{sec:conclusion}

This paper proposes MuPPET, a novel low-precision CNN training scheme that combines the use of fixed-point and floating-point representations to produce a network trained for FP32 inference.  
By introducing a precision-switching mechanism that decides at run time an appropriate transition point between different {\tb precision regimes}, the proposed framework achieves Top-1 validation accuracies comparable to that achieved by state-of-the-art FP32 training regimes while delivering significant speedup in terms of training time. Quantitative evaluation demonstrates that MuPPET's training strategy generalises across CNN architectures and datasets by {\tb adapting} the training process to the target CNN-dataset pair during run time. 
Overall, MuPPET enables the utilisation of the low-precision hardware units available on modern specialised processors, 
such as next-generation GPUs, FPGAs and TPUs, to yield improvements in training time and energy efficiency without impacting the resulting accuracy.
Future work will focus on applying the proposed framework to the training of LSTMs, where the training process is more sensitive to gradient quantisation \cite{approximate_LSTM}, as well as on the extension of MuPPET to include batch size and learning rate as part of its hyperparameters.  
Furthermore, we will explore improved quantisation techniques that could enable training convergence for bitwidths even lower than 8-bit fixed-point.

\section{Acknowledgements}
The support of the EPSRC Centre for Doctoral Training in High Performance Embedded and Distributed Systems (HiPEDS, Grant Reference EP/L016796/1) is gratefully acknowledged.

\bibliography{references}

\begin{thebibliography}{34}
\providecommand{\natexlab}[1]{#1}
\providecommand{\url}[1]{\texttt{#1}}
\expandafter\ifx\csname urlstyle\endcsname\relax
  \providecommand{\doi}[1]{doi: #1}\else
  \providecommand{\doi}{doi: \begingroup \urlstyle{rm}\Url}\fi

\bibitem[Alistarh et~al.(2017)Alistarh, Grubic, Li, Tomioka, and
  Vojnovic]{Dan2017}
Alistarh, D., Grubic, D., Li, J., Tomioka, R., and Vojnovic, M.
\newblock {QSGD: Communication-Efficient SGD via Gradient Quantization and
  Encoding}.
\newblock In \emph{Advances in Neural Information Processing Systems
  (NeurIPS)}, pp.\  1709--1720. 2017.

\bibitem[{Burgess}(2019)]{nvidia_turing_2019}
{Burgess}, J.
\newblock {RTX ON – The NVIDIA TURING GPU}.
\newblock In \emph{IEEE Hot Chips 31 Symposium (HCS)}, pp.\  1--27, 2019.

\bibitem[Cai et~al.(2018)Cai, Chen, Zhang, Yu, and Wang]{efficient_arch_search}
Cai, H., Chen, T., Zhang, W., Yu, Y., and Wang, J.
\newblock {Efficient Architecture Search by Network Transformation}.
\newblock In \emph{AAAI Conference on Artificial Intelligence}, 2018.

\bibitem[Chen et~al.(2017)Chen, Hu, Xu, Zhou, Zhou, and Xu]{chen2017fxpnet}
Chen, X., Hu, X., Xu, N., Zhou, H., Zhou, H., and Xu, N.
\newblock {FxpNet: Training deep convolutional neural network in fixed-point
  representation}.
\newblock In \emph{International Joint Conference on Neural Networks (IJCNN
  2017)}, May 2017.

\bibitem[Chen et~al.(2019)Chen, Chen, Han, and Hu]{moDNN}
Chen, X., Chen, D.~Z., Han, Y., and Hu, X.~S.
\newblock {moDNN: Memory Optimal Deep Neural Network Training on Graphics
  Processing Units}.
\newblock \emph{IEEE Transactions on Parallel and Distributed Systems (TPDS)},
  30\penalty0 (3):\penalty0 646–661, 2019.

\bibitem[Coleman et~al.(2019)Coleman, Kang, Narayanan, Nardi, Zhao, Zhang,
  Bailis, Olukotun, R{\'e}, and Zaharia]{time_to_accuracy_2019}
Coleman, C., Kang, D., Narayanan, D., Nardi, L., Zhao, T., Zhang, J., Bailis,
  P., Olukotun, K., R{\'e}, C., and Zaharia, M.
\newblock {Analysis of DAWNBench, a Time-to-Accuracy Machine Learning
  Performance Benchmark}.
\newblock \emph{SIGOPS Oper. Syst. Rev.}, 53\penalty0 (1):\penalty0 14--25,
  2019.

\bibitem[Courbariaux et~al.(2015)Courbariaux, Bengio, and
  David]{courbariaux2015training}
Courbariaux, M., Bengio, Y., and David, J.-P.
\newblock {Training deep neural networks with low precision multiplications}.
\newblock In \emph{Workshop in International Conference in Learning
  Representations}, 2015.

\bibitem[De~Sa et~al.(2017)De~Sa, Feldman, R{\'e}, and
  Olukotun]{asyncsgd2017isca}
De~Sa, C., Feldman, M., R{\'e}, C., and Olukotun, K.
\newblock {Understanding and Optimizing Asynchronous Low-Precision Stochastic
  Gradient Descent}.
\newblock In \emph{International Symposium on Computer Architecture (ISCA)},
  pp.\  561--574, 2017.

\bibitem[De~Sa et~al.(2015)De~Sa, Zhang, Olukotun, R\'{e}, and R\'{e}]{Sa2015}
De~Sa, C.~M., Zhang, C., Olukotun, K., R\'{e}, C., and R\'{e}, C.
\newblock {Taming the Wild: A Unified Analysis of Hogwild-Style Algorithms}.
\newblock In \emph{Advances in Neural Information Processing Systems
  (NeurIPS)}, pp.\  2674--2682. 2015.

\bibitem[Deng et~al.(2009)Deng, Dong, Socher, Li, Li, and
  Fei-Fei]{deng2009imagenet}
Deng, J., Dong, W., Socher, R., Li, L.-J., Li, K., and Fei-Fei, L.
\newblock {ImageNet: A Large-Scale Hierarchical Image Database}.
\newblock In \emph{IEEE Conference on Computer Vision and Pattern Recognition
  (CVPR)}, pp.\  248--255, 2009.

\bibitem[Devarakonda et~al.(2017)Devarakonda, Naumov, and Garland]{AdaBatch}
Devarakonda, A., Naumov, M., and Garland, M.
\newblock {AdaBatch: Adaptive Batch Sizes for Training Deep Neural Networks}.
\newblock \emph{arXiv}, 2017.

\bibitem[{Fowers} et~al.(2018){Fowers}, {Ovtcharov}, {Papamichael},
  {Massengill}, {Liu}, {Lo}, {Alkalay}, {Haselman}, {Adams}, {Ghandi}, {Heil},
  {Patel}, {Sapek}, {Weisz}, {Woods}, {Lanka}, {Reinhardt}, {Caulfield},
  {Chung}, and {Burger}]{brainwave_2018}
{Fowers}, J., {Ovtcharov}, K., {Papamichael}, M., {Massengill}, T., {Liu}, M.,
  {Lo}, D., {Alkalay}, S., {Haselman}, M., {Adams}, L., {Ghandi}, M., {Heil},
  S., {Patel}, P., {Sapek}, A., {Weisz}, G., {Woods}, L., {Lanka}, S.,
  {Reinhardt}, S.~K., {Caulfield}, A.~M., {Chung}, E.~S., and {Burger}, D.
\newblock {A Configurable Cloud-Scale DNN Processor for Real-Time AI}.
\newblock In \emph{International Symposium on Computer Architecture (ISCA)},
  pp.\  1--14, 2018.

\bibitem[Gan et~al.(2015)Gan, Wang, Yang, Yeung, and Hauptmann]{Gan2015}
Gan, C., Wang, N., Yang, Y., Yeung, D.-Y., and Hauptmann, A.~G.
\newblock {DevNet: A Deep Event Network for Multimedia Event Detection and
  Evidence Recounting}.
\newblock In \emph{IEEE Conference on Computer Vision and Pattern Recognition
  (CVPR)}, pp.\  2568--2577, 2015.

\bibitem[Gupta et~al.(2015)Gupta, Agrawal, Gopalakrishnan, and
  Narayanan]{Gupta_2015}
Gupta, S., Agrawal, A., Gopalakrishnan, K., and Narayanan, P.
\newblock {Deep Learning with Limited Numerical Precision}.
\newblock In \emph{32nd International Conference on Machine Learning (ICML)},
  pp.\  1737--1746, 2015.

\bibitem[He et~al.(2018)He, Gkioxari, Dollar, and Girshick]{He_2018}
He, K., Gkioxari, G., Dollar, P., and Girshick, R.
\newblock {Mask R-CNN}.
\newblock \emph{IEEE Transactions on Pattern Analysis and Machine Intelligence
  (TPAMI)}, 2018.

\bibitem[Huang et~al.(2017)Huang, Liu, van~der Maaten, and
  Weinberger]{huang2017densely}
Huang, G., Liu, Z., van~der Maaten, L., and Weinberger, K.~Q.
\newblock {Densely Connected Convolutional Networks}.
\newblock In \emph{IEEE Conference on Computer Vision and Pattern Recognition
  (CVPR)}, pp.\  2261--2269, 2017.

\bibitem[Johnson \& Guestrin(2018)Johnson and Guestrin]{RAIS}
Johnson, T.~B. and Guestrin, C.
\newblock {Training Deep Models Faster with Robust, Approximate Importance
  Sampling}.
\newblock In \emph{Advances in Neural Information Processing Systems 31}, pp.\
  7265--7275. 2018.

\bibitem[Jouppi et~al.(2017)]{Jouppi2017}
Jouppi, N.~P. et~al.
\newblock {In-Datacenter Performance Analysis of a Tensor Processing Unit}.
\newblock In \emph{44th Annual International Symposium on Computer Architecture
  (ISCA)}, pp.\  1--12, 2017.

\bibitem[Kerr et~al.(2018)Kerr, Merril, Demouth, and Tran]{cutlass}
Kerr, A., Merril, D., Demouth, J., and Tran, J.
\newblock {CUTLASS: Fast Linear Algebra in CUDA C}, Sep 2018.
\newblock URL \url{https://devblogs.nvidia.com/cutlass-linear-algebra-cuda/}.

\bibitem[{Kouris} \& {Bouganis}(2018){Kouris} and
  {Bouganis}]{kouris_drone2018iros}
{Kouris}, A. and {Bouganis}, C.
\newblock {Learning to Fly by MySelf: A Self-Supervised CNN-Based Approach for
  Autonomous Navigation}.
\newblock In \emph{2018 IEEE/RSJ International Conference on Intelligent Robots
  and Systems (IROS)}, pp.\  1--9, 2018.

\bibitem[{Kouris} et~al.(2018){Kouris}, {Venieris}, and
  {Bouganis}]{kouris2018cascadecnn}
{Kouris}, A., {Venieris}, S.~I., and {Bouganis}, C.
\newblock {CascadeCNN: Pushing the Performance Limits of Quantisation in
  Convolutional Neural Networks}.
\newblock In \emph{2018 28th International Conference on Field Programmable
  Logic and Applications (FPL)}, pp.\  155--1557, 2018.

\bibitem[Lin et~al.(2014)Lin, Maire, Belongie, Hays, Perona, Ramanan,
  Doll{\'a}r, and Zitnick]{coco2014}
Lin, T.-Y., Maire, M., Belongie, S., Hays, J., Perona, P., Ramanan, D.,
  Doll{\'a}r, P., and Zitnick, C.~L.
\newblock {Microsoft COCO: Common Objects in Context}.
\newblock In \emph{European Conference on Computer Vision (ECCV)}, pp.\
  740--755, 2014.

\bibitem[Loquercio et~al.(2018)Loquercio, Maqueda, del Blanco, and
  Scaramuzza]{Loquercio_2018}
Loquercio, A., Maqueda, A.~I., del Blanco, C.~R., and Scaramuzza, D.
\newblock {DroNet: Learning to Fly by Driving}.
\newblock \emph{IEEE Robotics and Automation Letters}, 3\penalty0 (2):\penalty0
  1088--1095, 2018.

\bibitem[Micikevicius et~al.(2018)Micikevicius, Narang, Alben, Diamos, Elsen,
  Garcia, Ginsburg, Houston, Kuchaiev, Venkatesh, and
  Wu]{micikevicius2018mixed}
Micikevicius, P., Narang, S., Alben, J., Diamos, G., Elsen, E., Garcia, D.,
  Ginsburg, B., Houston, M., Kuchaiev, O., Venkatesh, G., and Wu, H.
\newblock {Mixed Precision Training}.
\newblock In \emph{International Conference on Learning Representations
  (ICLR)}, 2018.

\bibitem[Migdalas et~al.(2013)Migdalas, Pardalos, and
  V{\"a}rbrand]{migdalas2013multilevel}
Migdalas, A., Pardalos, P.~M., and V{\"a}rbrand, P.
\newblock \emph{{Multilevel Optimization: Algorithms and Applications}},
  volume~20.
\newblock Springer Science \& Business Media, 2013.

\bibitem[Peng et~al.(2019)Peng, Li, and Wang]{typicality_sampling}
Peng, X., Li, L., and Wang, F.-Y.
\newblock {Accelerating Minibatch Stochastic Gradient Descent using Typicality
  Sampling}.
\newblock \emph{arXiv:1903.04192 [cs, math, stat]}, 2019.

\bibitem[Rhu et~al.(2016)Rhu, Gimelshein, Clemons, Zulfiqar, and Keckler]{vDNN}
Rhu, M., Gimelshein, N., Clemons, J., Zulfiqar, A., and Keckler, S.~W.
\newblock {vDNN: Virtualized Deep Neural Networks for Scalable,
  Memory-efficient Neural Network Design}.
\newblock In \emph{IEEE/ACM International Symposium on Microarchitecture
  (MICRO)}, pp.\  18:1--18:13, 2016.

\bibitem[Rizakis et~al.(2018)Rizakis, Venieris, Kouris, and
  Bouganis]{approximate_LSTM}
Rizakis, M., Venieris, S.~I., Kouris, A., and Bouganis, C.
\newblock {Approximate FPGA-based LSTMs under Computation Time Constraints}.
\newblock In \emph{14 International Symposium on Applied Reconfigurable
  Computing (ARC)}, 2018.

\bibitem[{Sze} et~al.(2017){Sze}, {Chen}, {Yang}, and {Emer}]{LowPrecSurvey}
{Sze}, V., {Chen}, Y., {Yang}, T., and {Emer}, J.~S.
\newblock {Efficient Processing of Deep Neural Networks: A Tutorial and
  Survey}.
\newblock \emph{Proceedings of the IEEE}, 105\penalty0 (12):\penalty0
  2295--2329, 2017.

\bibitem[Szegedy et~al.(2017)Szegedy, Ioffe, Vanhoucke, and
  Alemi]{Szegedy_2017}
Szegedy, C., Ioffe, S., Vanhoucke, V., and Alemi, A.
\newblock {Inception-v4, Inception-ResNet and the Impact of Residual
  Connections on Learning}.
\newblock In \emph{AAAI Conference on Artificial Intelligence}, 2017.

\bibitem[Wang et~al.(2018)Wang, Choi, Brand, Chen, and
  Gopalakrishnan]{8bitFPTrain}
Wang, N., Choi, J., Brand, D., Chen, C.-Y., and Gopalakrishnan, K.
\newblock {Training Deep Neural Networks with 8-bit Floating Point Numbers}.
\newblock In \emph{Advances in Neural Information Processing Systems
  (NeurIPS)}, pp.\  7675--7684. 2018.

\bibitem[Yin et~al.(2018)Yin, Pananjady, Lam, Papailiopoulos, Ramchandran, and
  Bartlett]{gradient_diversity}
Yin, D., Pananjady, A., Lam, M., Papailiopoulos, D., Ramchandran, K., and
  Bartlett, P.
\newblock {Gradient Diversity: a Key Ingredient for Scalable Distributed
  Learning}.
\newblock In \emph{21st International Conference on Artificial Intelligence and
  Statistics (AISTATS)}, pp.\  1998--2007, 2018.

\bibitem[{Zhong} et~al.(2018){Zhong}, {Yan}, {Wu}, {Shao}, and
  {Liu}]{block-wise_arch_search}
{Zhong}, Z., {Yan}, J., {Wu}, W., {Shao}, J., and {Liu}, C.
\newblock {Practical Block-Wise Neural Network Architecture Generation}.
\newblock In \emph{IEEE Conference on Computer Vision and Pattern Recognition
  (CVPR)}, pp.\  2423--2432, 2018.

\bibitem[Zhou et~al.(2016)Zhou, Ni, Zhou, Wen, Wu, and Zou]{Zhou_16}
Zhou, S., Ni, Z., Zhou, X., Wen, H., Wu, Y., and Zou, Y.
\newblock {DoReFa-Net: Training Low Bitwidth Convolutional Neural Networks with
  Low Bitwidth Gradients}.
\newblock \emph{CoRR}, 2016.

\end{thebibliography}
\bibliographystyle{icml2020}

\appendix
\onecolumn
\section{Appendix A}
\label{sec:AppendixA}
This section contains the larger versions of all the figures in the paper for enhanced clarity.
\begin{figure}[h]
    \centering
    \includegraphics[trim={6cm 30 9cm 10}, clip, width = \textwidth]{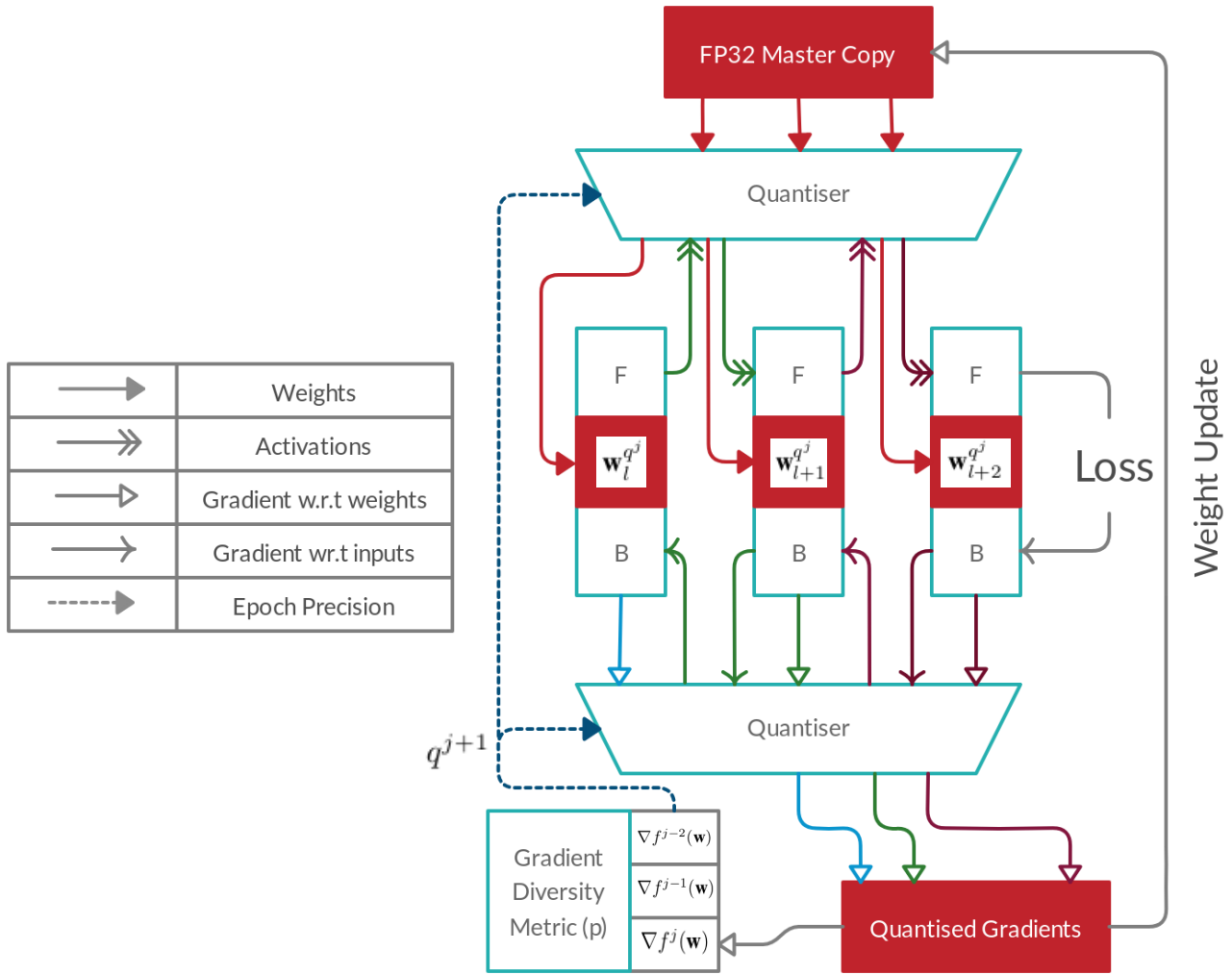}
    \caption{Precision-switching training scheme of MuPPET.}
\end{figure}

\begin{figure}[h]
    \centering
    \includegraphics[trim=0 0 1cm 0,clip,width = \textwidth]{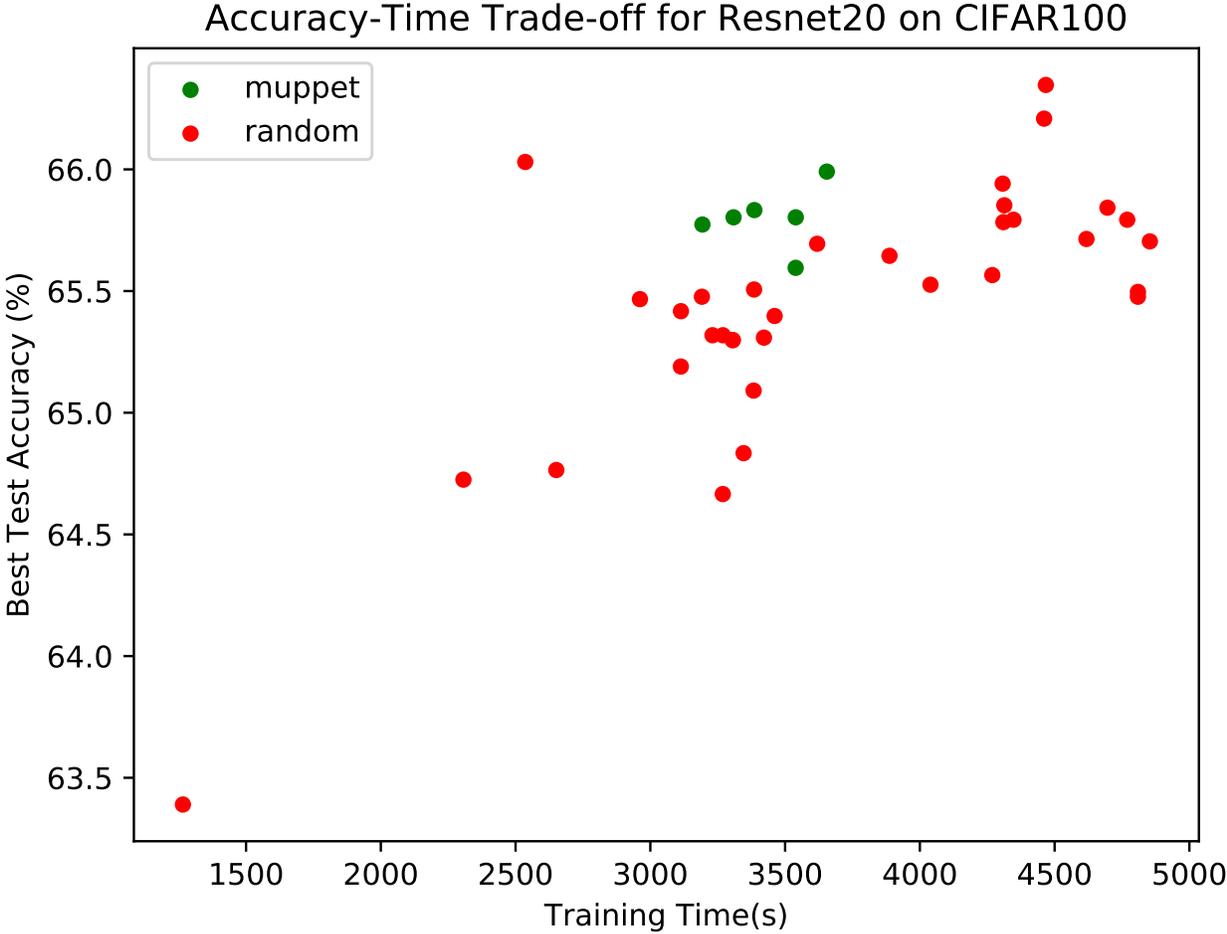}
    \caption{Accuracy vs time tradeoff for ResNet20 MuPPET runs on CIFAR100 - Final Test Accuracy}
\end{figure}

\begin{figure}[h]
    \centering
    \includegraphics[width = \textwidth]{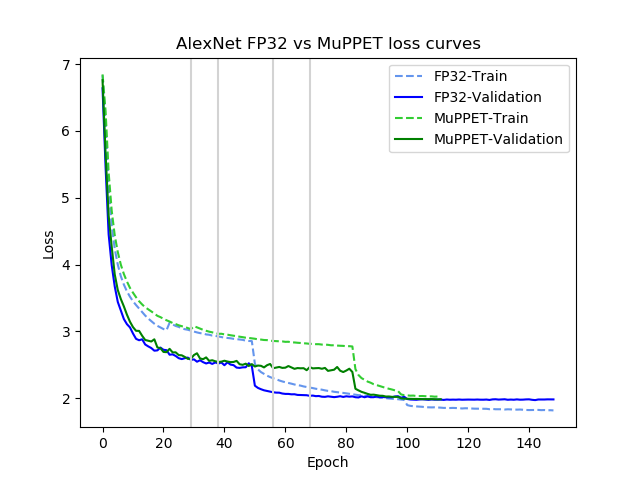}
    \caption{AlexNet training and validation loss values for FP32 and MuPPET on ImageNet.}
\end{figure}

\begin{figure}[h]
    \centering
    \includegraphics[width = \textwidth]{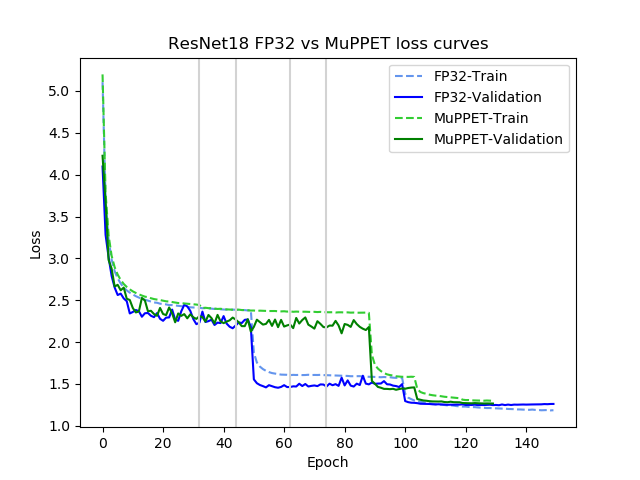}
    \caption{ResNet18 training and validation loss values for FP32 and MuPPET on ImageNet.}
\end{figure}

\begin{figure}[h]
    \centering
    \includegraphics[width = \textwidth]{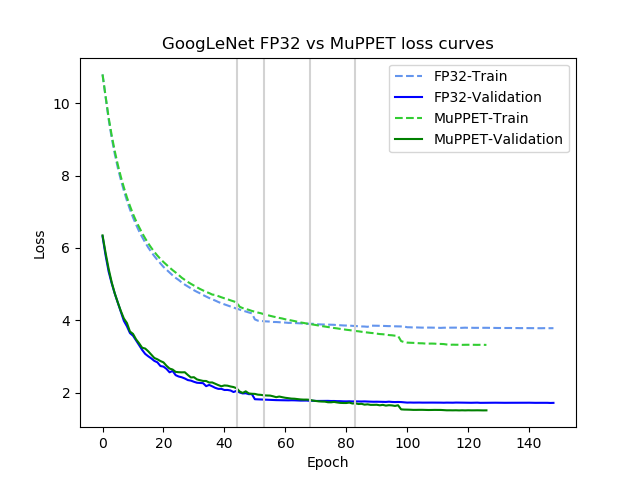}
    \caption{GoogLeNet training and validation loss values for FP32 and MuPPET on ImageNet.}
\end{figure}

\begin{figure}[h]
    \centering
    \includegraphics[trim=2 2 30 2.5,clip,width = \textwidth]{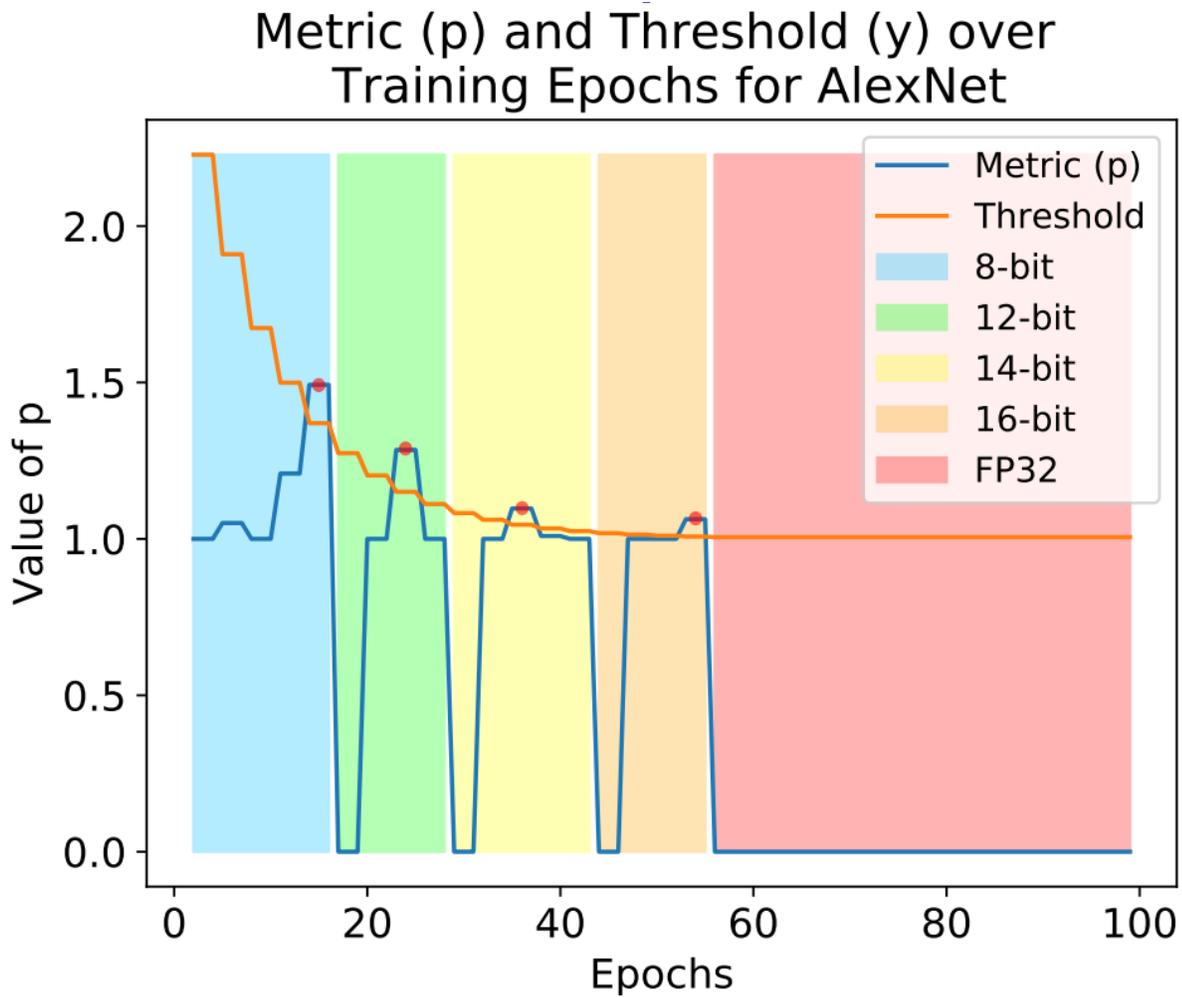}
    \caption{AlexNet CIFAR10 - Demonstration of the generalisability of $p$ over networks, datasets and epochs.}
\end{figure}

\begin{figure}[h]
    \centering
    \includegraphics[trim=5 5 5 5,clip,width = \textwidth]{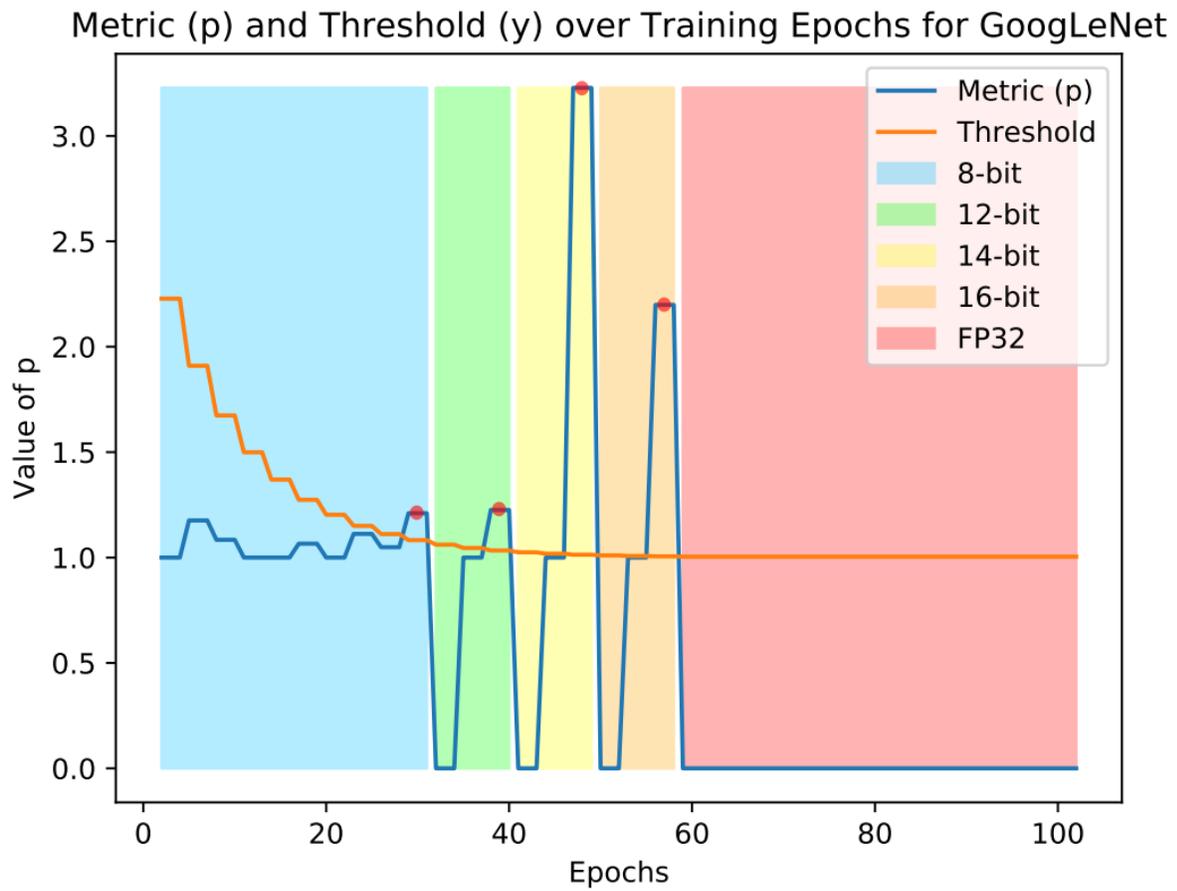}
    \caption{GoogLeNet CIFAR10 - Demonstration of the generalisability of $p$ over networks, datasets and epochs.}
\end{figure}

\begin{figure}[h]
    \centering
    \includegraphics[trim=5 5 0 5,clip,width = \textwidth]{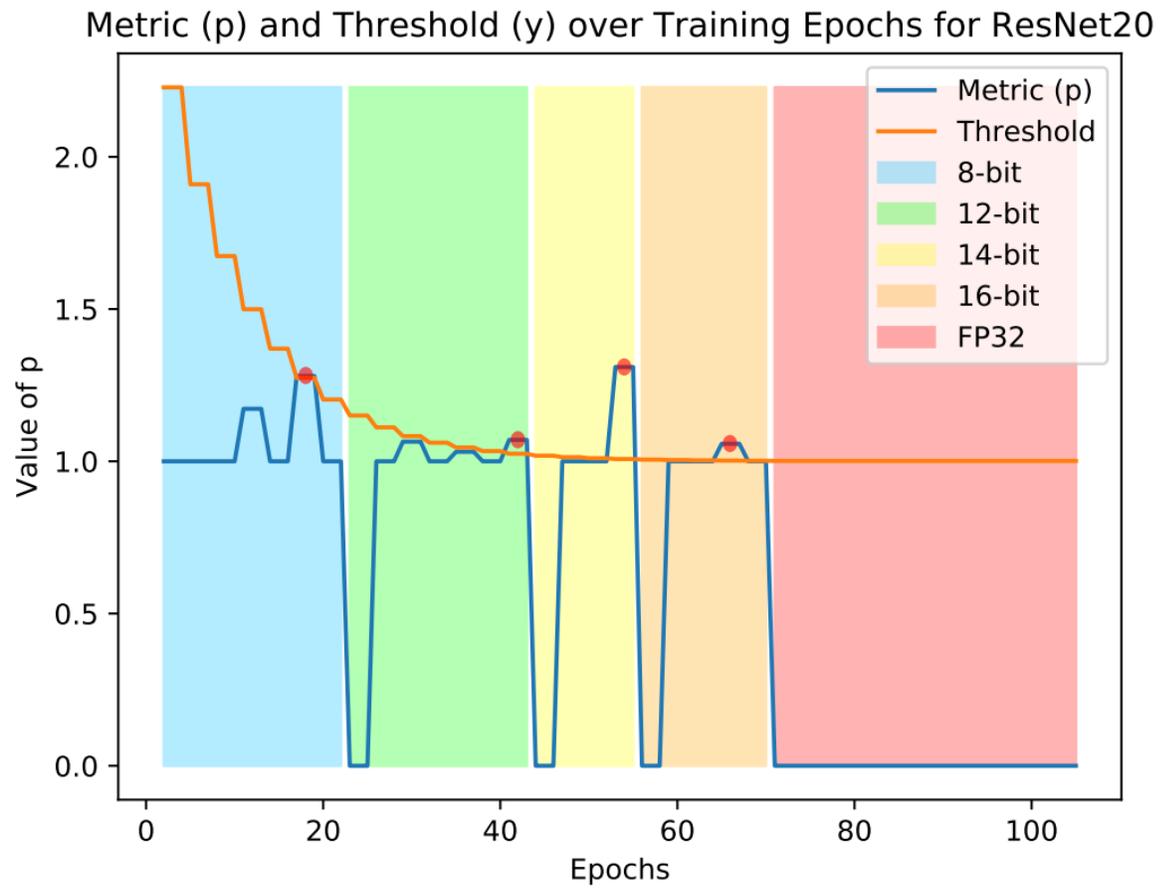}
    \caption{ResNet20 CIFAR10 - Demonstration of the generalisability of $p$ over networks, datasets and epochs.}
\end{figure}

\begin{figure}[h]
    \centering
    \includegraphics[trim=0 0 5 5,clip,width = \textwidth]{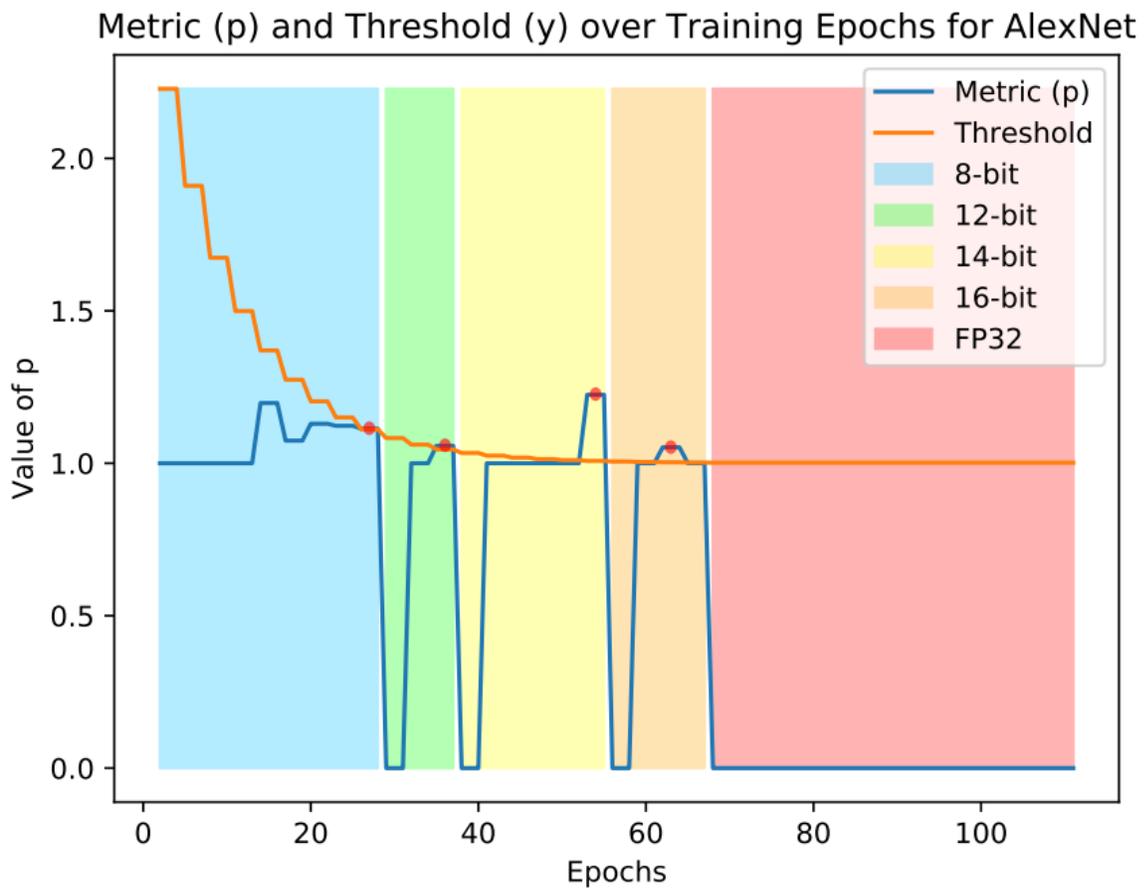}
    \caption{AlexNet ImageNet - Demonstration of the generalisability of $p$ over networks, datasets and epochs.}
\end{figure}

\begin{figure}[h]
    \centering
    \includegraphics[trim=5 0 0 5,clip,width = \textwidth]{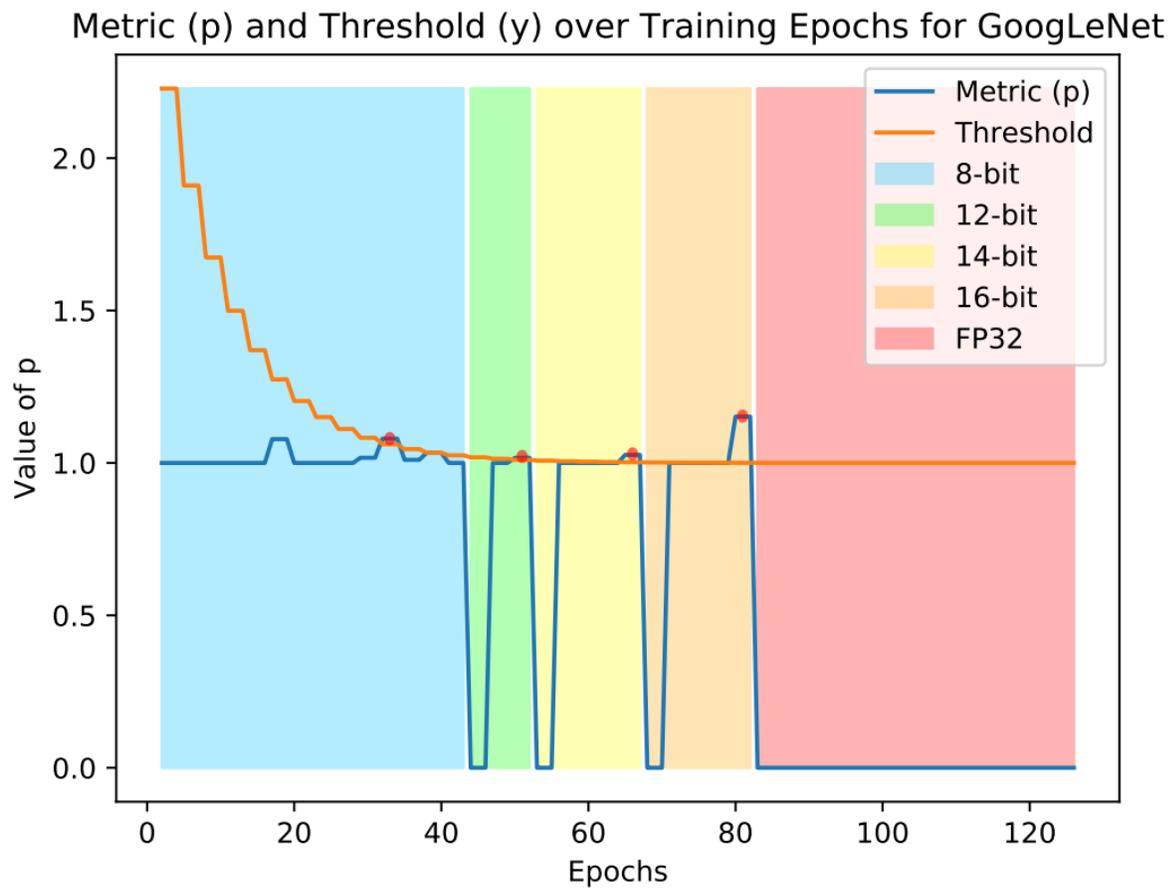}
    \caption{GoogLeNet ImageNet - Demonstration of the generalisability of $p$ over networks, datasets and epochs.}
\end{figure}

\begin{figure}[h]
    \centering
    \includegraphics[trim=0 0 5 0,clip,width = \textwidth]{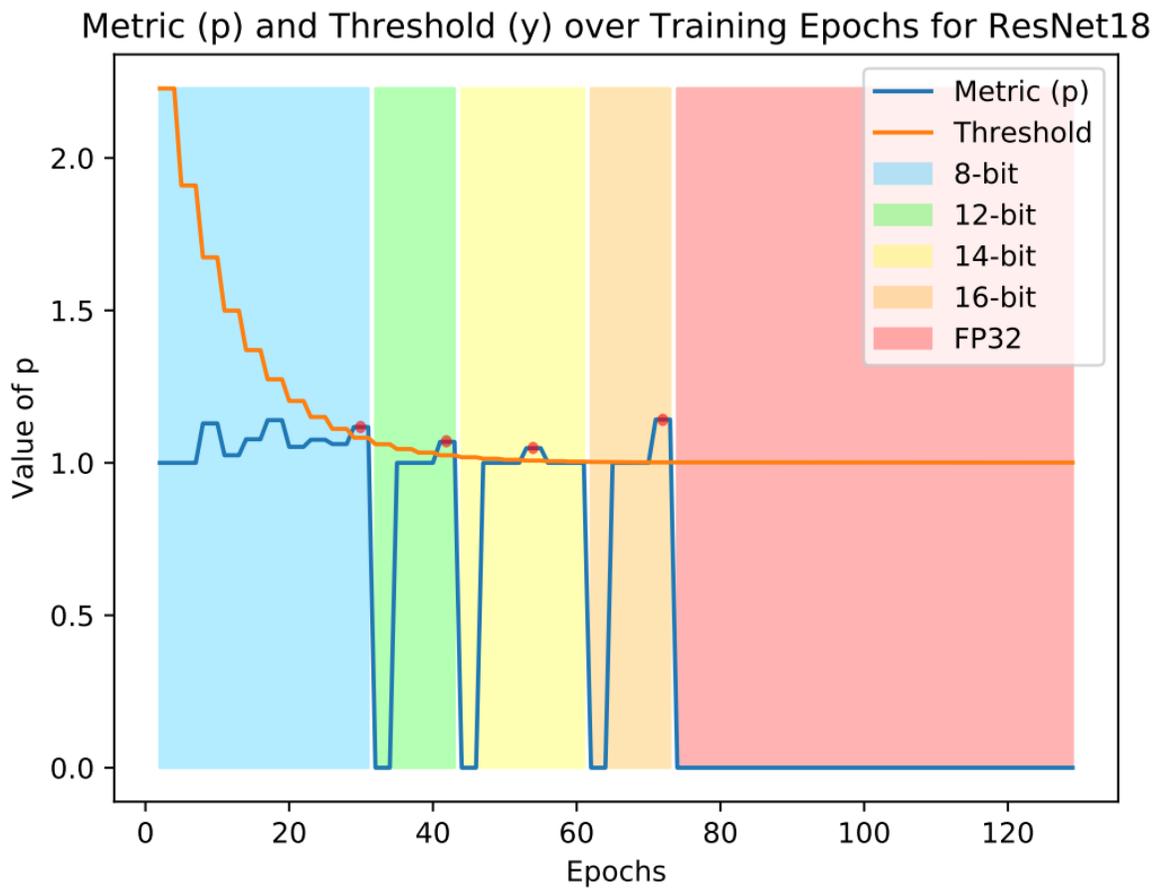}
    \caption{ResNet18 ImageNet - Demonstration of the generalisability of $p$ over networks, datasets and epochs.}
\end{figure}

\end{document}